\documentclass[opre]{informs3} %

\usepackage{setspace}
\usepackage{dsfont}
\usepackage{multirow}
\OneAndAHalfSpacedXII 

\usepackage{natbib}
 \bibpunct[, ]{(}{)}{,}{a}{}{,}%
 \def\bibfont{\small}%

\theoremstyle{example}
\newtheorem{example}{Example}
\usepackage[normalem]{ulem}

\usepackage{hyperref}
\usepackage{mathabx}
\usepackage{enumitem}
\usepackage{makecell}
\usepackage[ruled,vlined]{algorithm2e}
\usepackage{algpseudocode}

\EquationsNumberedThrough    

\begin{document}




\TITLE{Collaborative Intelligence in Sequential Experiments: A Human-in-the-Loop Framework for Drug Discovery}

\ARTICLEAUTHORS{%
\AUTHOR{Jinghai He}
\AFF{Department of Industrial Engineering and Operations Research, University of California Berkeley, \EMAIL{jinghai\_he@berkeley.edu}}
\AUTHOR{Cheng Hua}
\AFF{Antai College of Economics and Management, Shanghai Jiao Tong University, \EMAIL{cheng.hua@sjtu.edu.cn}}
\AUTHOR{Yingfei Wang}
\AFF{Foster School of Business, University of Washington, \EMAIL{yingfei@uw.edu}}
\AUTHOR{Zeyu Zheng}
\AFF{Department of Industrial Engineering and Operations Research, University of California Berkeley, \EMAIL{zyzheng@berkeley.edu}}
} 

\ABSTRACT{%
Drug discovery is a complex process that involves sequentially screening and examining a vast array of molecules to identify those with the target properties. This process, also referred to as sequential experimentation, faces challenges due to the vast search space, the rarity of target molecules, and constraints imposed by limited data and experimental budgets. To address these challenges, we introduce a human-in-the-loop framework for sequential experiments in drug discovery. This collaborative approach combines human expert knowledge with deep learning algorithms, enhancing the discovery of target molecules within a specified experimental budget. The proposed algorithm processes experimental data to recommend both promising molecules and those that could improve its performance to human experts. Human experts retain the final decision-making authority  based on these recommendations and their domain expertise, including the ability to override algorithmic recommendations.  We applied our method to drug discovery tasks using real-world data and found that it consistently outperforms all baseline methods, including those which rely solely on human or algorithmic input. This demonstrates the complementarity between human experts and the algorithm. Our results provide key insights into the levels of humans' domain knowledge, the importance of meta-knowledge, and effective work delegation strategies. Our findings suggest that such a framework can significantly accelerate the development of new vaccines and drugs by leveraging the best of both human and artificial intelligence.}%
\HISTORY{}

\maketitle

%

\section{Introduction}

Drug discovery is arguably one of the most expensive and riskiest processes in the world \citep{lou2021ai}. It can take up to ten years and \$2–3 billion to bring a new medicine to market, with a success rate as low as 1\% \citep{das2021acceleratedanti}. In addition, the cost of developing new drugs has been steadily increasing. Since the 1950s, the number of new drugs approved per \$1 billion of research and development investment has halved approximately every nine years, a trend consistent over the past seven decades. This phenomenon is known as the anti-Moore's law of the pharmaceutical industry, or Eroom's Law \citep{ringel2020breakingeroom}. In this work, we aim to describe a framework that integrates human expertise with deep learning algorithms to enhance the screening of effective molecules, a crucial step in drug discovery.

Drug discovery processes involve sequentially screening a large pool of candidate molecules. 
This step, known as sequential experimentation, aims to identify target molecules with desired properties. Sequential experimentation serves as an important, yet costly, step before the discovered molecules proceed to clinical trials and await potential approval by the Food and Drug Administration (FDA). New molecules are examined in sequential batches, and the decision-maker needs to decide which molecules to test in each batch. The main challenges faced by sequential experiments include the vast search space, the scarce presence of target molecules, and the limited availability of relevant data and opportunities for experiments.




The simplest approach in sequential experimentation is to randomly select molecules for evaluation using advanced high-throughput lab technologies. While this method is straightforward and ensures unbiased sampling, it is highly time-consuming and resource-intensive, with a low probability of successfully identifying effective molecules due to the vast number of possibilities \citep{powell2012optimal}.
Alternative methods include testing molecules related to already discovered target molecules. This strategy has a higher initial success rate and offers a more focused approach, but it suffers from limited exploration, potentially overlooking effective molecules that differ structurally from known targets. This can lead to inefficiency in the long term \citep{ravina2011evolution}. Relying on expert knowledge to screen molecules recommended by domain experts still maintains a central role in drug discovery \citep{sturm2021coordinating}. It allows for informed decision-making and customization of molecular designs. However, it is subject to expert bias and the potential overlooking of novel, unconventional molecules that fall outside current expert understanding. Lastly, data-driven algorithms could assist in identifying potential candidates efficiently and adapt over time with exposure to more data. However, their performance heavily depends on the availability and quality of training data. In cases of insufficient or biased data, these algorithms risk poor performance, leading to ineffective or misleading results \citep{cai2020transfer}. 


To address these challenges, we make an attempt to present a \textit{human-in-the-loop} framework for sequential experiments in drug discovery that aims to provide an effective method for sequential experimentation through collaborative decisions made by both deep learning algorithms and human experts. These algorithms are trained on data that is progressively acquired through experiments. They provide recommendations for molecules, identifying those with beneficial properties or those that can enhance algorithmic performance. Human experts utilize their domain knowledge to assess and, if necessary, override these algorithmic recommendations, playing a key role in refining the algorithm's performance. We anticipate that humans and deep learning algorithms, when working together, can exhibit complementarity (or complementary performance) in this challenging task and outperform the performance of either learning algorithms or humans working independently, as is widely discussed in the information systems literature \citep{fugener2021will, van2021machine, fugener2022cognitive}.

The proposed framework employs algorithms powered by state-of-the-art transformer models that offer evolutionary-scale atomic-level molecule predictions \citep{lin2023evolutionary}, and Bayesian neural networks that yield predictions of properties and their associated uncertainties, while the human experts are pharmaceutical chemists. The proposed method balances exploration, aimed at enhancing algorithm performance for subsequent rounds, and exploitation, which focuses on the discovery of target molecules. In human's collaborative work, meta-knowledge -- which refers to individuals' awareness and understanding of their own knowledge and abilities \citep{fugener2022cognitive} -- is essential because it facilitates more effective role assignment and conflict resolution. This insight drives the design of our proposed algorithm, particularly in the use of Bayesian neural networks and the uncertainty score. The model and data uncertainties, detailed in \S \ref{sec:uncertainty_quant}, provide compelling evidence for human experts to discern which recommendations to follow. Based on these estimated uncertainties and predictions, the algorithm not only recommends promising molecules to the experts but also suggests molecules that could aid in its own improvement. Human experts will make the molecule-selection decisions based on the algorithm's outputs and their domain knowledge, which enables them to override the algorithm's recommendations if necessary. The outcomes from each round are also used to enhance the algorithm's performance for the subsequent screening round.

The rationale behind our framework design can be interpreted in two-fold: 1) From a \textit{performance perspective}, we facilitate collaboration between humans and algorithms, capitalizing on each's strengths in drug discovery tasks. While the algorithm excels in computational efficiency and generalizability, capable of learning from data \citep{van2021machine, chen2024philosopher}, it may face challenges due to insufficient data, potentially leading to noticeable errors and a decreased ability to learn effectively.
Human experts possess private information that can enhance performance \citep{balakrishnan2022improving}, particularly in challenging tasks where algorithms might under-perform \citep{fugener2022cognitive}. Specifically, in our context, while they are less efficient in managing large candidate pools, they excel at providing valuable suggestions and identifying errors in the algorithms' outputs using their domain-specific knowledge. 2) From a \textit{behavioral perspective}, \cite{dietvorst2018overcoming} show that people are more likely to use imperfect algorithms if they can modify them, even slightly. Despite the algorithm's imperfections due to data limitations, particularly in the initial stages, humans can amend the algorithm's results, thereby enhancing its performance. This strategy strengthens the practical synergy between humans and algorithms and achieves complementarity \citep{donahue2022human}.

We evaluate our proposed human-in-the-loop framework through a series of experiments conducted on a real-world therapeutic peptide dataset for the discovery of therapeutic peptides satisfying multiple
properties.  Our framework enhances the efficiency of sequential experiments in drug discovery. Empirical results provide evidence for the superiority of our approach and demonstrate the effectiveness of each design component. Notably,  within a given budget, our method at least doubles the efficiency of identifying
promising molecules compared to all baseline methods that do not include human participation and only human experts, exhibiting complementarity between
human experts and the deep learning algorithm, which can aid in discovering new
drugs in practical scenarios and has the potential to rapidly identify vaccines or drugs
during virus outbreaks.

Our study  provides important insights for collaborative intelligence in sequential experiments, while existing literature on human-AI collaboration mainly focuses on
supervised learning tasks. In our context, approaches that rely solely on human input or exclusively on algorithmic processes are less effective, while a complementarity between human experts and algorithms can be realized in challenging tasks. Additionally, the effectiveness of collaboration improves with the  expertise of the human participants. Moreover, our findings highlight the importance of meta-knowledge \citep{fugener2022cognitive} in enhancing human-algorithm collaboration. Furthermore, our analysis reveals that the strategy of delegation within collaborative intelligence significantly impacts overall success and, that in our context of sequential experimentation, delegating work from algorithms to humans is more effective than the reverse, from humans to algorithms.

\section{Related Literature} 
\label{sec: Related works}


In recent years, statistical and computational methods have accelerated drug discovery, known as virtual or in silico screening in the medical field \citep{negoescu2011knowledge}. 
While many researchers focus on accurate prediction of molecular properties, such as chemical and biochemical properties, through machine learning and deep learning algorithms \citep{zoffmann2019machine, ekins2019exploiting, vamathevan2019applications}, the process of drug design requires the selection of molecules for synthesis and testing in each round, which is a sequential decision problem.

\subsection{Sequential Experiments}

One stream of studies models sequential experiments as a ranking and selection problem \citep{hong2021review}, whose goal is to determine the best alternative given the information obtained from a sequence of experiments. 
In this stream, a seminal paper by \cite{negoescu2011knowledge} applies knowledge gradient, a Bayesian optimization approach, to search for the best compound and dramatically reduce the number of tests required. This technique has also been used to search for targeted regions of RNA molecules \citep{li2018knowledge} and to identify the right dose in clinical trials \citep{nasrollahzadeh2022dynamic}.

Our setting differs from that described in \cite{negoescu2011knowledge} in several ways. First, while their paper focuses on optimizing molecules with a single property, we consider molecules that satisfy multiple properties. Second, our search space includes all possible molecules within a specified length, whereas they focus on modifications of a base molecule, a scope several orders of magnitude larger. Third, and we regard this as the primary difference, concerns the objective: we aim to maximize the number of target molecules identified that meet a set of requirements. In contrast, they seek to identify the best molecule. This approach is more aligned with the drug discovery 
practice, as the target molecules identified still need to undergo several phases of clinical trials to evaluate their effects in humans. The success rate of being finally approved is very low, and discovering more candidate drug molecules can increase the success rate of drug discovery \citep{reda2020machine}.



Another stream of research formulates sequential experimental design for drug discovery as an active learning problem \citep{reker2015active}. Active learning allows the algorithm to actively direct the training process on how to obtain new data points for training the model  \citep{zheng2006selectively, saar2007decision, deodhar2017active, li2022more}. In this way, a viable algorithm can be trained with limited available data. This makes it particularly suitable for situations where unlabeled data is abundant but manual labeling is expensive, such as drug discovery \citep{schneider2018automating}. Another stream of work focuses on contextual online learning; see \cite{miao2022context}, \cite{chen2022primal}, \cite{keskin2022data}, \cite{li2023dynamic}, \cite{zhang2024contextual}, among others. Their focus is usually on understanding the trade-off with regard to regret order, which has a different goal and algorithm design from our work. 

A limitation of this stream of approach is that the basic goal of active learning is to improve the model performance rather than to find target molecules by sequential experiments, which deviates from the ultimate goal of drug discovery. Contrary to the existing literature, we integrate the concept of active learning with optimization in sequential experiments to maximize the discovery of target molecules using limited available data.

\subsection{Human-Algorithm Collaboration}

Our approach also relates to intelligence augmentation in human and algorithm connections. Intelligence augmentation leverages the combined strengths of humans and algorithms to achieve superior performance than either could attain independently \citep{jain2021editorial,wang2023human}. By fostering a collaborative partnership between human experts and algorithms, our method embodies the principles of intelligence augmentation.
The work by \cite{fugener2021will} explores the merits and pitfalls of AI collaboration, noting that while AI can potentially boost performance, it may also diminish the synergy of knowledge between humans and AI. Specifically, it could erode the unique contributions of human insight. Similarly, \cite{donahue2022human} delves into the complementary of human-algorithm collaboration. Other studies discuss the problem of routing instances to either the human or the algorithm \citep{okati2021differentiable,de2021classification}. \cite{straitouri2023improving}  examines a classification task, in which the algorithm suggests a selection of possible labels from which the human makes the final decision. \cite{cabitza2021studying} explores various approaches for combining human and algorithmic predictions. In \cite{wilder2020learning} a machine takes on the tasks of deciding which instances require human input and then fusing
machine and human judgments.\cite{fugener2022cognitive} investigates the dynamics of delegation within human-AI teams, finding that human performance can decline due to inadequate meta-knowledge, which refers to an individual's ability to accurately evaluate their own skills, leading to suboptimal delegation decisions, a task algorithms perform better at. In our study, we demonstrate that in the context of drug discovery, the algorithm's ability to recognize its own limitations, as expressed through model and data uncertainties estimated by the model, is crucial. This meta-knowledge is essential for optimizing collaboration and achieving better outcomes. 

The interplay between human decision-making and algorithms has attracted great interest across various fields. This includes notable studies on their application in financial services \citep{sinha2008incorporating, ge2021human}, the application of AI in healthcare settings \citep{zhou2019tumor, lebovitz2021ai, jussupow2021augmenting, zhou2023spoiled}, the identification of the optimal exoskeleton assistance to humans \citep{zhang2017human}, and in customer service sectors \citep{schanke2021estimating}. Additionally, studies by \cite{sun2022predicting} and \cite{kawaguchi2021will} explore human adherence to and collaboration with algorithms through field experiments in the logistics and retail sectors. In the domain of drug discovery, \cite{lou2021ai} highlights the potential of AI and medical expertise by reviewing practices across various pharmaceutical companies. Data-driven algorithms have emerged as effective tools for molecule prediction \citep{adjeroh2018feature, jumper2021alphafold} and drug discovery \citep{negoescu2011knowledge, vamathevan2019applications,  mouchlis2021advances}. However, current algorithm-assisted drug optimization suffers from data hungriness in real-world drug discovery tasks \citep{cai2020transfer}, considering there is only limited availability of labeled data for new drug development.

A seminal study by \cite{dietvorst2018overcoming} examines the behavioral effects of human-algorithm interaction, especially when individuals are allowed to adjust algorithmic outputs. Additionally, \cite{balakrishnan2022improving} suggests that integrating unique human insights, which are inaccessible to algorithms, can enhance performance. Excluding human input from decision-making processes removes the opportunity to leverage such invaluable insights. Our framework, in conjunction with these existing findings, leverages the specialized domain knowledge of pharmaceutical experts, which is crucial for enhancing performance metrics, and allows them to adjust algorithm outputs.

Nevertheless, most literature differs from our setting in that they consider supervised learning tasks and study human-algorithm collaboration for regression or classification tasks. In comparison,  we focus on sequential experimentation where the algorithm continues to learn and interact with the human decision-maker and the environment, receiving partial feedback (i.e. bandit feedback) of only the chosen option but not the others. While the settings and reward models are difference from ours, some work has looked more specifically at human-algorithm collaboration in
bandit settings. \cite{gao2021human}  learns from batched historical human data to develop a routing algorithm
that assigns each task to either an algorithm or a human, under bandit feedback. 
In \cite{chan2019assistive},  a robot assists a human in a
bandit setting to maximize cumulative reward.  The robot makes the final decisions that overrides the human's recommendation. However, 
the robot only observes which arms the human pulls but not the reward
associated with each pull.

\subsection{Meta-Knowledge and Uncertainty Quantification} \label{sec: meta-knowledge}

Our research is closely aligned with the concept of meta-knowledge, which is defined as the ability to assess one's own capabilities—specifically, to understand what one knows. This ability is critical across various domains \citep{evans2011metaknowledge}. The seminal work by \cite{fugener2022cognitive} explores meta-knowledge within the context of human-AI collaboration, particularly in decision-making and task delegation. They suggest that parties lacking certainty should delegate tasks to the other party, utilizing meta-knowledge to guide this process. Additionally, they observed that while humans often lack sufficient meta-knowledge, AI typically exhibits a high level of meta-knowledge. However, when AI is applied without careful consideration, it can produce overconfident predictions, leading to suboptimal outcomes \citep{wilson2020bayesian}. This issue is often rooted in uncertainties—either stemming from the data itself (data uncertainty) or from incomplete model knowledge (model uncertainty). Notably, standard neural networks in AI lack mechanisms to provide certainty estimates \citep{gawlikowski2023survey}.

Recognizing the importance of meta-knowledge in AI, researchers emphasize \textit{uncertainty quantification} as a method to equip AI algorithms with the ability to self-assess their predictions. This technique not only boosts the confidence in AI-generated predictions but also enhances collaboration with human experts by providing clear, quantifiable insights into decision-making processes. Pioneering works, such as those by \cite{hullermeier2021aleatoric}, have laid the foundation for distinguishing between aleatoric (data-related) and epistemic (model-related) uncertainties in neural networks, offering strategies to effectively model and quantify them. With growing interest in this field, researchers continue to develop robust methods for uncertainty estimation in neural networks, highlighted by contributions from \cite{gal2016dropout}, \cite{lakshminarayanan2017simple}, \cite{blundell2015weight},  \cite{van2020uncertainty} and \cite{zheng2023doubly}. For broader concepts of uncertainty quantification and a deeper dive into its applications in neural networks, comprehensive reviews are available from \cite{ghanem2017handbook} and \cite{gawlikowski2023survey}.

\section{Model of Sequential Experiments} \label{sec: process of Drug Discovery}


The sequential experiment in drug discovery aims to identify molecules that satisfy all $K$ property requirements in $R$ rounds, under a fixed experimental budget $B$ per round. This budget is typically constrained by factors such as money, time, and available space for experimentation. In this work, we focus on molecules that can be expressed as a vector $\boldsymbol{x} = \left(x_1,x_2,\cdots,x_l\right)$, where $l\in \{1,2,\cdots,L\}$ is the length of the molecule and $L$ represents the maximum length considered. Examples of such molecules include peptides, proteins, nucleic acids (such as DNA or RNA), macromolecules, polymers, and others.  Each element $x_i\in \mathcal{V}$ is a component of the molecule, where $\mathcal{V}$ is the set of all considered components. We define $C = |\mathcal{V}|$ as the total number of these components. The search space $\mathcal{X}$ includes all potential molecules up to a maximum length of $L$. This space is defined as:
\begin{equation}
    \mathcal{X}=\left\{\boldsymbol{x} \mid \boldsymbol{x}= \left(x_1,x_2,\cdots,x_l\right), l\in \left\{1,2,\cdots,L\right\}, x_i \in \mathcal{V}\right\}{.}
\end{equation}
The total number of molecules in this space, or its cardinality, is $|\mathcal{X}|=\sum_{i=1}^L C^i=\frac{C^{L+1}-C}{C-1}$. 

We consider $K$ properties of candidate molecules, denoted by a vector $\boldsymbol{y} = \boldsymbol{f(x)} = \left(f_1(\boldsymbol{x}),f_2(\boldsymbol{x})\dots,f_K(\boldsymbol{x})\right)$. Each property $f_k(\boldsymbol{x})$ is either a real number representing a property measured by its numeric value, i.e., $f_k(\boldsymbol{x})\in\mathcal{R}$, or a binary value indicating the pass or fail of a requirement, i.e., $f_k(\boldsymbol{x})\in \{0,1\}$. The properties of each molecule $\boldsymbol{x}$ remain unknown until $\boldsymbol{x}$ is synthesized and tested in a laboratory experiment. Our goal is to identify molecules that meet all $K$ requirements. That is, for all $k\in \{1,2\cdots,K\},$
\begin{equation}
    f_k(\boldsymbol{x}) \geq h_k,
\end{equation}
where $h_k$ represents the threshold for property $k$. In general, $h_k$ can be a real number in a continuous domain, or $h_k=1$ for binary delete{test} outcomes. 
We refer to a molecule as a \textit{target molecule} if it meets all $K$ requirements. In drug discovery, identifying a target molecule is known as a \textit{hit}.

The search budget per laboratory round, denoted by $B$, represents the total number of molecules that can be synthesized and tested in each round. Practically, the duration of each round may vary from days to weeks or months, depending on the task at hand. The total number of rounds, denoted by $R$, is often constrained by time and financial resources. Consequently, the overall search budget throughout the discovery process is $B \times R$, indicating the cumulative number of molecules that can be synthesized and tested.

A sequential experiment search policy, denoted as $\pi$, determines the selection of $B$ molecules at the beginning of each round for synthesis and experimental testing. Let $S_{\pi}$ represent the total number of target molecules identified following policy $\pi$. We aim to develop a sequential experiment policy, $\pi^*$, that maximizes the hit rate. The hit rate is defined as the ratio of target molecules discovered to  the total number of $B \times R$ tests, expressed as:
\begin{equation}
    \pi^* = \argmax_{\pi} \frac{S_{\pi}}{B\times R}.
\end{equation}

\begin{example}
In therapeutic peptide drug discovery, each molecule $\boldsymbol{x}$ is a peptide, and each component $x_i\in\mathcal{V}$ is an amino acid that constitutes the peptide. There are a total of 20 amino acids globally ($C=|\mathcal{V}|=20$). Suppose we consider peptides comprising fewer than 25 amino acids ($L=25$) that satisfy $K=3$ criteria: anti-bacterial, non-toxic, and anti-microbial properties. An example of a target peptide is $\boldsymbol{x} = \left(\mathrm{G,F,R,K,F,H,K,F,W,A}\right)$, where capital letters denote different types of amino acids.  The 3 properties of this peptide are $\boldsymbol{y}= (1,1,1)$.
The search space for this problem is $\frac{20^{26} - 20}{19} \approx 3.53\times 10^{32}$. If in each round we test $B=100$ peptides over a total of $R=50$ rounds, this means we need to identify target molecules in only $5,000$ experimental tests from this vast search space ($\sim 10^{32}$). 
\end{example}

In practice, identifying target molecules that meet all $K$ criteria is challenging for four main reasons. (1) \textbf{Huge search space.} The size of the search space, $|\mathcal{X}| =\frac{C^{L+1}-C}{C-1}$, is immense. 
(2) \textbf{Few target molecules.} Only a tiny fraction ($< 1\%$) of molecules might fulfill even a single property requirement \citep{drews2000drug}, much less all $K$ property requirements simultaneously. (3) \textbf{Limited budget.} In the real-world drug development process, either the search budget per round ($B$) or the total number of rounds ($R$) is often limited. This constraint makes exploration difficult and the data collected scarce. (4) \textbf{Cold start.} At the onset of the search process, no data is available. Consequently, traditional data-driven methods (such as machine learning and deep learning) struggle due to a lack of sufficient training samples.

The scarcity of target molecules within a vast search space, coupled with a shortage of available training samples, impedes the effective use of data-driven algorithms for locating target molecules efficiently. However, human experts, equipped with domain knowledge, can steer the algorithm towards exploring potential molecules and steer clear of those that are evidently unlikely. In the following section, we introduce a human-in-the-loop framework for sequential experiments, emphasizing collaborative decision-making between the algorithm and human experts.
The notation in this paper is summarized in Table \ref{tab:variable}.

\begin{table}[htbp]
    \centering
    \footnotesize
    \caption{Notation Table}
    \begin{tabular}{cc|cc}
    \hline
        \thead{Notation}  & \thead{Explanation} & \thead{Notation}  & \thead{Explanation}\\ \hline
      $ \mathcal{V}$  &  \thead{Set of constituents \\that compose the molecule} &     $\boldsymbol{\mu}$ & \thead{Predicted mean vector}\\
      $\mathcal{X}$  &  \thead{Molecule search space} &     $\boldsymbol{\sigma}_m(\boldsymbol{x})\in \mathcal{R}^K$ & \thead{ Model uncertainty \\(epistemic uncertainty) }\\ 
      $\boldsymbol{x}\in \mathcal{X}$  &  \thead{Molecule sequence} &    $\boldsymbol{\sigma}_d(\boldsymbol{x})\in \mathcal{R}^K$ & \thead{Data uncertainty \\(aleatoric uncertainty) }\\ 
      $\boldsymbol{z} \in \mathcal{R}^{|\mathcal{Z}|}$ & \thead{Real-number embedding of a molecule} &        $\pi$ & \thead{Search policy}\\
      $K\in Z_+$     & \thead{Number of properties considered}  &     $S_\pi \in Z_+$ & \thead{Number of target molecules \\found under policy $\pi$}\\
      
      $\boldsymbol{y(x)} \in \mathcal{R}^K$ & \thead{Vector of values for $K$ properties} &      $S_\pi^t \in Z_+$ & \thead{Number of target molecules\\ found  under policy $\pi$ \\in the first $t$ rounds} \\
      
      $\mathcal{D}_\text{labeled}^{(t)}$,$\mathcal{D}_\text{unlabeled}^{(t)}$  &  \thead{Labeled/Unlabeled dataset in round $t$ } &     $B\in Z_+$ & \thead{Search budget in each round}\\
      
      $\mathcal{C}_{\theta}$ & \thead{Bayesian neural network model \\ parameterized by parameters $\theta$} &      $R\in Z_+$ & \thead{Total number of experiment rounds}\\
    $r_{un}(\boldsymbol{x}) \in \mathcal{R}$ & \thead{Uncertainty score} & $r_{se}(\boldsymbol{x}) \in \mathcal{R}$ & \thead{Search score}\\
\hline
    \end{tabular}
    \label{tab:variable}
\end{table}

\section{Human-in-the-loop Sequential Experiment} 
\begin{figure}[htbp]
\FIGURE
{\includegraphics[scale=0.35]{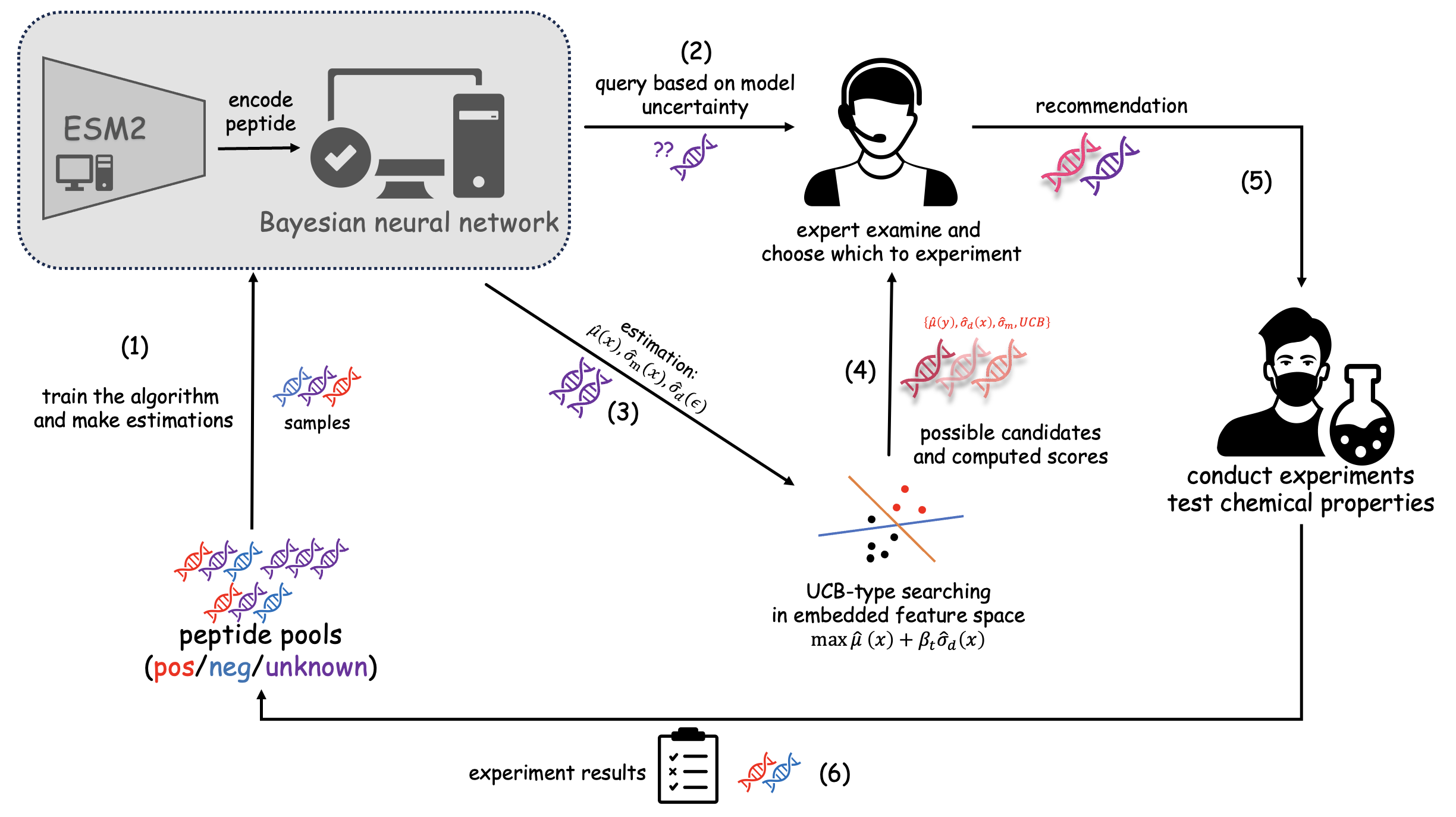}}
{Framework of Human-in-the-Loop Sequential Experiment in Drug Discovery.\label{fig: human in the loop framework}}
{Red molecule sequences represent positive samples, blue samples indicate negative ones, and purple samples denote those without labels.
In step (1), we first encode peptide sequences into a numerical representation with the ESM2 model introduced in \S \ref{sec:ESM2}, and train Bayesian neural networks using all labeled data, which then predicts the properties of unknown molecules and estimates their prediction uncertainty. During step (2), the molecules with the $q$ highest model uncertainties are recommended to the experts. In steps (3) and (4), the algorithm also recommends the molecule with the $h$ highest search score. Subsequently, in step (5), the expert evaluates all molecules recommended in steps (2) and (4), selecting $B$ molecules based on the algorithm's scoring and their domain knowledge. The experimenter then conducts laboratory experiments on these selected $B$ molecules and updates the labeled dataset with the results in step (6). This updated dataset is utilized for training in the next iteration.
}
\end{figure}

Framework \label{sec: human-in-loop}
In this section, we introduce the human-in-the-loop sequential experiment framework. 
The entire process spans $R$ rounds; during each round, the algorithm and human experts collaborate to decide a set of molecules for experimental testing. The algorithm employs a deep learning transformer model for molecule embedding and Bayesian neural network models for property predictions and uncertainty estimation. The framework is illustrated in Figure \ref{fig: human in the loop framework}.  

The interaction between humans and algorithms occurs in two primary ways:  1) The algorithm suggests molecules based on two key criteria: their superior predicted properties or their prediction uncertainties, providing both prediction and uncertainty scores as meta-knowledge to human experts to aid in decision-making. 2) Human experts, applying their domain knowledge, either approve or override the algorithm's recommendations, thereby concluding the final molecule selection decision. This decision finalizes the molecule choices for the current round and also provides these molecules, along with their tested results, as new inputs for the model, influencing the algorithm's performance in future rounds.

\subsection{An Overview of the Algorithm}

In each round $t$, the molecules in the search pool fall into two groups: the labeled dataset $\mathcal{D}^{(t)}_\text{labeled}$ and the unlabeled dataset $\mathcal{D}^{(t)}_\text{unlabeled}$. The labeled dataset comprises molecules that have been previously tested in the lab, with known labels $\boldsymbol{y} = \boldsymbol{f(x)}$, while the unlabeled dataset encompasses the extensive pool of all molecules yet to be tested. The datasets are defined as follows:
$$
\mathcal{D}^{(t)}_\text{labeled}=\left\{\left(\boldsymbol{x}_1, \boldsymbol{y}_1\right), \left(\boldsymbol{x}_2, \boldsymbol{y}_2\right) \ldots,\left(\boldsymbol{x}_{N^{(t)}}, \boldsymbol{y}_{N^{(t)}}\right)\right\}, \quad \mathcal{D}^{(t)}_\text{unlabeled}=\left\{\boldsymbol{x}_{N^{(t)}+1}, \ldots, \boldsymbol{x}_J\right\},
$$
where $N^{(t)}=(t-1)B$ represents the number of molecules synthesized and tested before round $t$, and $J$ is the total number of molecules in the search pool\footnote{The search pool could either be the complete pool, where $J=|\mathcal{X}| =\frac{C^{L+1}-C}{C-1}$, or a specific subset defined by human experts. The first approach is termed molecule generation, also known as de novo design, while the latter approach is referred to as molecule picking, involving screening from a predefined molecule pool.}.



We employ a deep learning transformer model to encode molecules into a real-number embedding feature space $\mathcal{Z}$, as detailed in \S \ref{sec:ESM2}. 
Subsequently, using the encoding and label for each molecule in $\mathcal{D}^{(t)}_\text{labeled}$, we update the Bayesian neural network model for each property, as detailed in \S \ref{sec:bayesian_NN}. The algorithm then predicts $K$ properties for each unlabeled molecule $\boldsymbol{x}_i^u \in \mathcal{D}^{(t)}_\text{unlabeled}$, providing: (i) an estimated mean value vector $\boldsymbol{\mu}(\boldsymbol{x}_i^u)$; (ii) an estimated model uncertainty vector $\boldsymbol{\sigma}_m(\boldsymbol{x}_i^u)$; (iii) an estimated data uncertainty vector $\boldsymbol{\sigma}_d(\boldsymbol{x}_i^u)$, with further details in \S \ref{sec:uncertainty_quant}.

The algorithm utilizes these estimates, which constitute its meta-knowledge, to calculate two distinct scores: the uncertainty score $r_{un}(\boldsymbol{x}^u)$ and the search score $r_{se}(\boldsymbol{x}^u)$. These scores serve two different purposes: one for enhancing algorithm performance and the other for optimizing molecule search. Based on these scores, the algorithm then recommends two separate batches of molecules from $\mathcal{D}_\text{unlabeled}$. The first batch comprises $q$ molecules $\{\boldsymbol{x}_{un,i}^{(t)}\}_{i=1}^q$ with the highest uncertainty scores $r_{un}$. 
These molecules are selected to potentially improve model performance, with further details in \S \ref{sec:active details}. 
The second batch includes $h$ molecules $\{\boldsymbol{x}_{se, i}^{(t)}\}_{i=1}^h$ with the top search scores $r_{se}$, which are considered likely candidates for target molecules based on the current algorithm's predictions, as detailed in \S \ref{sec:UCB}.


Human experts are tasked with selecting $B$ molecules $\{\boldsymbol{x}_{i}^{(t)}\}_{i=1}^B$ from the pool of molecules $\{\boldsymbol{x}_{se, i}^{(t)}\}_{i=1}^h\bigcup\{\boldsymbol{x}_{un,i}^{(t)}\}_{i=1}^q$ recommended by the algorithm for laboratory experiments. 
This decision-making process requires the experts to integrate algorithmic outputs with their domain knowledge. The specific role and contributions of the human experts in this process are detailed in \S \ref{sec: human}. 

The selected $B$ molecules will then be synthesized and tested for their properties in the laboratory. The results, including these molecules and their tested properties, will be incorporated into the labeled dataset for the next round: $\mathcal{D}_{\text{labeled}}^{(t+1)}=\mathcal{D}_{\text{labeled}}^{(t)}\bigcup\{(\boldsymbol{x}_{i}^{(t)},\boldsymbol{y}_{i}^{(t)})\}_{i=1}^B$. Our human-in-the-loop framework for sequential experiments is shown in Algorithm \ref{alg:HIL}.

\,

\begin{algorithm}[H]
    \caption{Human-in-the-loop Framework for Sequential Experiments}
    \label{alg:HIL}
    \SetAlgoLined
     \KwIn{ Initialize $K$ Bayesian neural network models $\{\mathcal{C}_{\theta_i^{(1)}}\}_{i=1}^K$,  $\mathcal{D}_{\text{unlabeled}}^{(1)}=\left\{ \boldsymbol{x}_j^u\right\}^J_{j=1}$, $\mathcal{D}_{\text{labeled}}^{(1)}=\emptyset$}
    \For{$t \in \{1,2,\cdots,R\}$}{1: Estimate $\{\boldsymbol{\mu}(\boldsymbol{x}_j^u),\boldsymbol{\sigma}_m(\boldsymbol{x}_j^u),\boldsymbol{\sigma}_d(\boldsymbol{x}_j^u)\}$ by the algorithm $\{\mathcal{C}_{\theta_i^{(t)}}\}_{i=1}^K$; \\
        2: Compute uncertainty scores $r_{un}(\boldsymbol{x}_j^u)$ and search scores $r_{se}(\boldsymbol{x}_j^u)$ for all $\boldsymbol{x}_j^u\in \mathcal{D}_{\text{unlabeled}}^{(t)}$;\\
    3: Recommend $h$ molecules $\{\boldsymbol{x}_{se, i}^{(t)}\}_{i=1}^h$  with the highest search scores, and $q$ molecules $\{\boldsymbol{x}_{un,i}^{(t)}\}_{i=1}^q$ with the highest uncertainty scores;\\
    4: Human experts examine the recommended molecules $\{\boldsymbol{x}_{un,i}^{(t)}\}_{i=1}^q\bigcup
    \{\boldsymbol{x}_{se, i}^{(t)}\}_{i=1}^h$  and choose $B$ molecules $\{\boldsymbol{x}_{i}^{(t)}\}_{i=1}^B$ for laboratory experiments, and obtain $\{\boldsymbol{y}_{i}^{(t)}\}_{i=1}^B$;\\
    5: Update $\mathcal{D}_{\text{labeled}}^{(t+1)}=\mathcal{D}_{\text{labeled}}^{(t)}\bigcup\{(\boldsymbol{x}_{i}^{(t)},\boldsymbol{y}_{i}^{(t)})\}_{i=1}^B$,$\mathcal{D}_{\text{unlabeled}}^{(t+1)}=\mathcal{D}_{\text{unlabeled}}^{(t)} \setminus \{\boldsymbol{x}_{i}^{(t)}\}_{i=1}^B$, and update $\theta_i^{(t)}$ with $\mathcal{D}_{\text{labeled}}^{(t+1)}$ to obtain the improved algorithm $\{\mathcal{C}_{\theta_i^{(t+1)}}\}_{i=1}^K$.
    }
\end{algorithm}

\subsection{Evolutionary-Scale Deep Learning Transformer Language Model}\label{sec:ESM2}

To analyze peptide sequences, which are essentially unstructured letter sequences, we employ an approach that converts them into structured representations. Recognizing the similarity between the limited types of amino acids in proteins and the 26 letters in the alphabet, we adopt deep natural language processing techniques. Specifically, we use the ESM2 model \citep{lin2023evolutionary}, a transformer-based protein language model that effectively converts protein sequences from character formats into a structured latent space.

\begin{figure}[htbp]
\FIGURE
{ \includegraphics[scale=0.78]{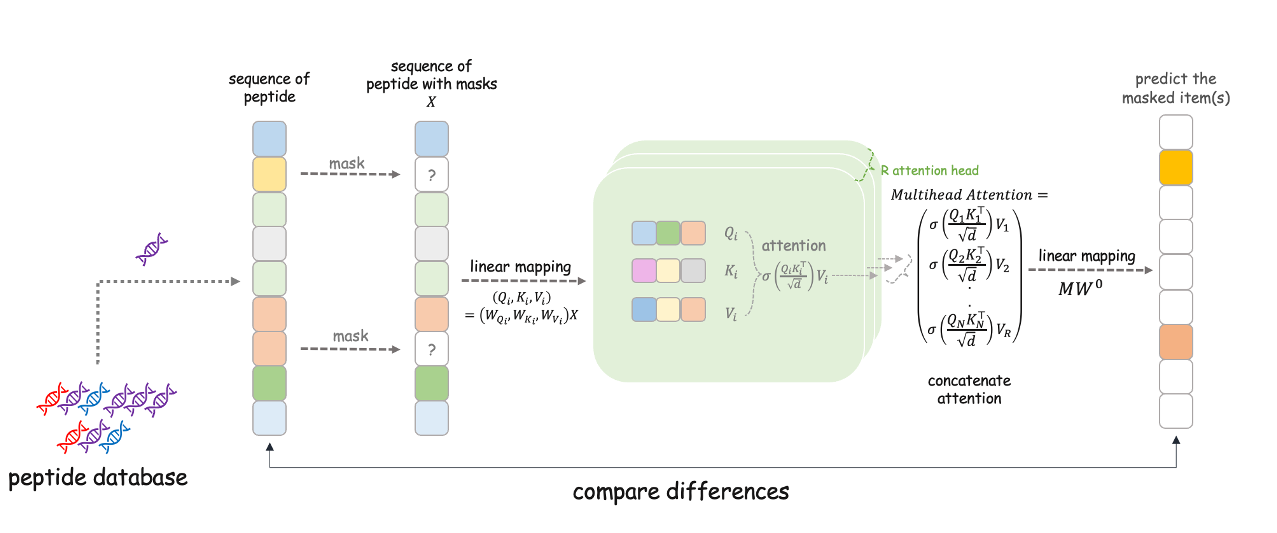}}
{Diagram of the ESM2 Transformer Architecture.\label{fig:ESM2}}
{The ESM2 model is a transformer-based deep learning language model specialized in protein sequence analysis. It transforms unstructured protein sequences into structured, real-number latent spaces. Utilizing an attention mechanism and a masking scheme, ESM2 predicts masked amino acids within sequences, providing high-dimensional embeddings.}
\end{figure}

The ESM2 model, trained on over 65 million existing peptide sequences, employs an attention mechanism and a masking scheme (Figure \ref{fig:ESM2}). This involves randomly masking an amino acid in a sequence and training the model to predict the masked amino acid. The model optimization focuses on a masked language model objective, where the loss function is defined as the negative log likelihood of correctly predicting the masked amino acid given the context of the unmasked sequence:
\begin{equation}
\mathcal{L}_{M L M}=\mathbb{E}_{x \sim X} \mathbb{E}_M \sum_{i \in M}-\log p\left(x_i \mid x_{/ M}\right), \label{eq:attention}
\end{equation}
where $x_i$ is the true amino acid, $M$ indicates the set of masked component, and $x_{/ M}$ is the masked sequence.

The model's self-attention mechanism processes a sequence of vectors and computes interactions among all elements. It transforms an input sequence $\left(x_1, \ldots, x_n\right)$ into an output sequence $\left(x_1^{\prime}, \ldots, x_n^{\prime}\right)$. The attention mechanism is based on scaled dot-product attention:
\begin{equation}
\text{Attention}_j=\sigma\left(\frac{Q_j K_j^T}{\sqrt{d}}\right) V_j, 
\end{equation}
where $Q_j$ (query), $K_j$ (key), and $V_j$ (value) are the $j^\text{th}$ linear transformations of the input sequence into matrices. The attention scores, obtained from the outer product of $Q$ and $K$, are scaled by the square root of the dimension of the matrices $d$ and transformed into a combination of the value sequence $V$ through the softmax function $\sigma$. The subscript $j\in\{1,2,\cdots,R\}$ represents the attention score for one from $R$ heads, which corresponds to one set of $Q$, $K$, and $V$ tuples. 

The use of multi-headed self-attention enables diverse inter-position interaction patterns. This is achieved as the concatenated output of multiple attention heads, represented by
\begin{equation}
\text{Multi-Attention}=(\text{Attention}_1, \ldots, \text{Attention}_R)
\end{equation}
is channeled into a subsequent feedforward layer, generating the latent representation of a sequence. The ESM2 model, with approximately 15 billion parameters and incorporating rotary position embedding, equips the model with the ability to express these protein molecules.

\begin{figure}[htbp]
    \FIGURE
    {\includegraphics[scale=0.23]{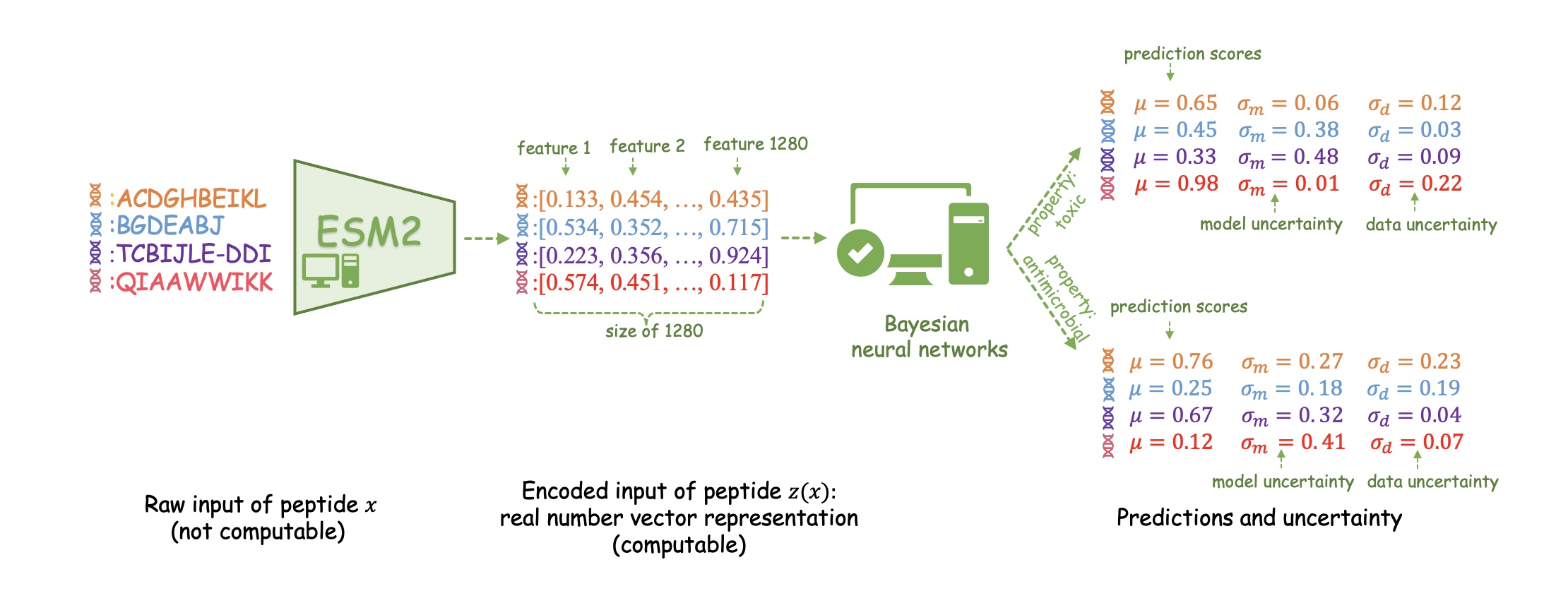}}
    {Encoding Peptide Molecules with ESM2 and Predicting with Bayesian Neural Networks. \label{fig:encoding}}
    {First, molecules are fed into the ESM2 transformer deep learning model to generate numerical encodings. These encodings are then input into Bayesian neural networks, which yield a mean, model uncertainty, and data uncertainty scores for each property. These scores generate uncertainty and search scores, serving as the foundational basis for the algorithm's recommendations.}
\end{figure}

In summary, we use the ESM2 model to encode peptide sequences $\boldsymbol{x} \in \mathcal{X}$ into $\boldsymbol{z}$ in a real-number embedding feature space $\mathcal{Z}$, where $|\mathcal{Z}|$ is the dimension of this space. This model ensures that the distance $\left\|\boldsymbol{z}_i-\boldsymbol{z}_j\right\|$ between embeddings of two peptides $\boldsymbol{x}_i$ and $\boldsymbol{x}_j$ is minimal for similar sequence patterns. The encoded peptide sequence $\boldsymbol{z}$ is thus represented as a tuple of features $\left(z_1, z_2, \ldots, z_{|\mathcal{Z}|}\right)$, serving as inputs for the Bayesian neural network models $\left\{\mathcal{C}_{\theta_j}\right\}_{j=1}^k$. The encoding process is illustrated in Figure \ref{fig:encoding}.

\subsection{Bayesian Neural Network}
\label{sec:bayesian_NN}



A Bayesian neural network is a neural network formulated within the Bayesian framework, where both the parameters and outputs are treated as random variables rather than fixed learned values. This type of neural network model is adept not only at making predictions but also at estimating the uncertainty associated with those predictions \citep{wilson2020bayesian}.

To simplify notation, let $X^{(t)}$ represent the set of encoded vectors $\boldsymbol{z}$ for all labeled molecules up to round $t$, and $Y^{(t)}$ denote their corresponding labels. Given the labeled dataset $\mathcal{D}^{(t)}_\text{labeled}=(X^{(t)}, Y^{(t)})$ in the $t^\text{th}$ round, the posterior distribution of the parameter $\theta$ is governed by Bayes' rule, which can be expressed as:
\begin{equation}
p(\theta \mid \mathcal{D}^{(t)}_\text{labeled})=\frac{p(Y^{(t)} \mid X^{(t)}, \theta) p(\theta)}{p(Y^{(t)} \mid X^{(t)})}, \label{eq:BNN_posterior}
\end{equation}
where $p(\theta)$ is the prior distribution, $p(Y^{(t)} \mid X^{(t)}, \theta)$ is the likelihood function, and $p(Y^{(t)} \mid X^{(t)})$ is the model evidence. More specifically, the model evidence is computed as $p\left(Y^{(t)} \mid X^{(t)}\right)=\int p\left(Y^{(t)} \mid X^{(t)}, \theta\right) p\left(\theta\right) d \theta $.

Consider $x$ as a new data input, representing the encoding of a molecule. The distribution of the corresponding label $y$ is derived using the following equation:
\begin{equation}
p\left(y \mid x, \mathcal{D}^{(t)}_\text{labeled}\right)=\int \underbrace{p\left(y \mid x, \theta\right)}_{\text {data uncertainty}} \underbrace{p(\theta \mid \mathcal{D}^{(t)}_\text{labeled})}_{\text {model uncertainty}} d \theta.
\end{equation}

However, direct calculation of \eqref{eq:BNN_posterior} poses significant computational challenges, particularly in computing the model evidence in the denominator. To circumvent this issue, we adopt variational inference \citep{graves2011practical} to approximate the posterior distribution. Within a specified distribution family, we optimize its parameters by minimizing the discrepancy between the variational distribution and the actual posterior distribution. This optimization process involves maximizing the evidence lower bound (ELBO) and employing the Kullback-Leibler divergence (KL divergence) to measure the difference between the two distributions.

Let $q_\omega(\theta)$ denote a variational distribution parameterized by $\omega$. The KL divergence between $q_\omega(\theta)$ and the true posterior $p(\theta \mid  \mathcal{D}^{(t)}_\text{labeled})$ is
$
    \mathrm{KL}\left(q \| p\right)=\int  q_\omega(\theta) \ln \frac{q_\omega(\theta)}{P(\theta \mid \mathcal{D}^{(t)}_\text{labeled})} d \theta,
$
where $q$ represents the variational distribution $q_\omega(\theta)$ and $p$ stands for the true posterior $p(\theta \mid  \mathcal{D}^{(t)}_\text{labeled})$. As the direct computation of the KL divergence involves the intractable true posterior, we optimize the evidence lower bound (ELBO) instead. The KL divergence can be decomposed into:
    $\mathrm{KL}(q \| p)=-\mathrm{ELBO} +\log p(\mathcal{Y}^{(t)} \mid \mathcal{X}^{(t)})$,
where 
$
\mathrm{ELBO}=\int  q_\omega(\theta)\ln \frac{p(\mathcal{Y}^{(t)} \mid \mathcal{X}^{(t)}, \theta)}{q_\omega(\theta)}d \theta.
$

An estimate of the posterior distribution $q_{\omega}(\theta)$ is obtained by optimizing the ELBO. This posterior is then utilized for prediction, as shown in the following equation:
\begin{equation}
    p(y \mid x, \mathcal{D}^{(t)}_\text{labeled})=\int p\left(y \mid x, \theta\right) q_{\omega}(\theta) d \theta. \label{eq_approx_posterior}
\end{equation}

Utilizing this predictive distribution, our approach diverges from conventional machine learning methods, which typically offer only a single point prediction of the mean value. Instead, we obtain a distribution of the potential outcomes. This distribution provides more than just a prediction of performance; it also offers estimations of uncertainty. These estimates are crucial for quantifying uncertainties \citep{gawlikowski2023survey}, which embody the model's meta-knowledge.

\subsection{Uncertainty Quantification} 
\label{sec:uncertainty_quant}

\begin{figure}[htbp]
\FIGURE
{\includegraphics[scale=0.8]{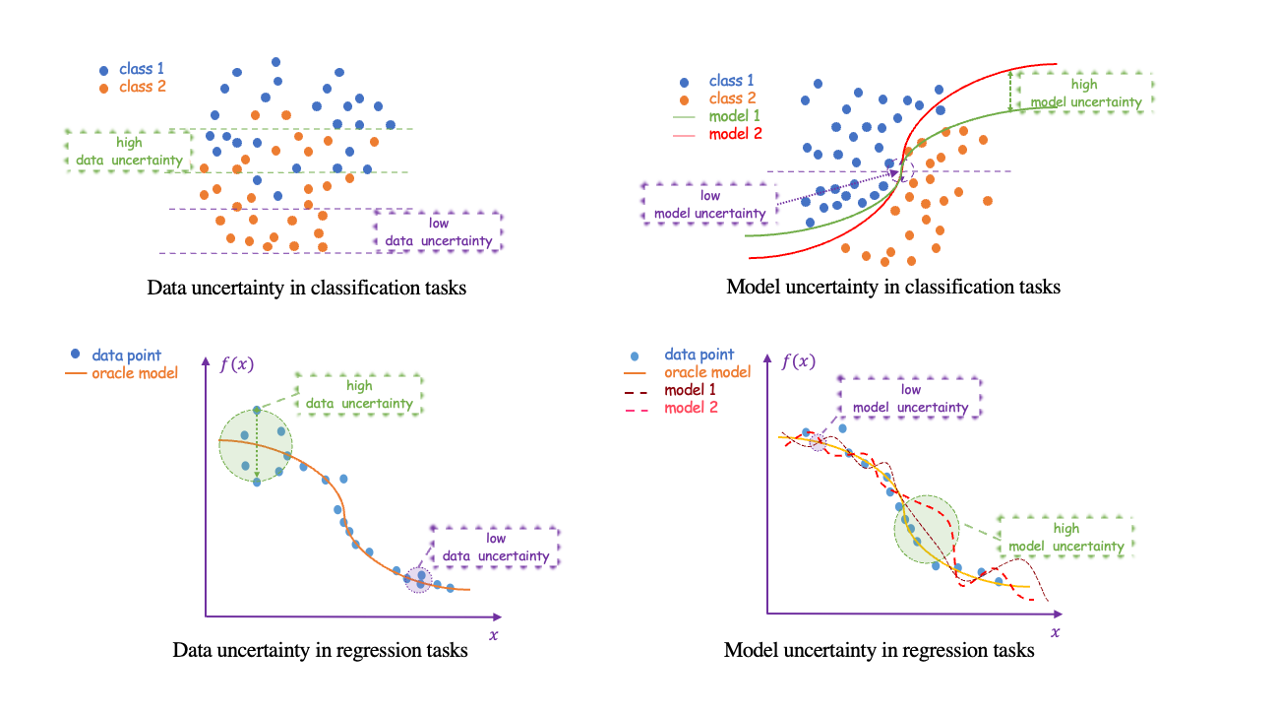}}
{Model Uncertainty and Data Uncertainty.\label{fig: uncertainty}}
{This figure illustrates the distinction between data uncertainty (aleatoric) and model uncertainty (epistemic) in both classification and regression tasks. Data uncertainty is intrinsic to the data itself and is irremovable. In contrast, model uncertainty stems from imperfections in the model, such as a lack of sufficient training data or an inadequate model structure, and can be mitigated through enhancements to the algorithm or the training process. 
}
\end{figure}

Estimating uncertainty becomes particularly vital in scenarios where the training dataset is limited in size. The estimated uncertainty can be leveraged to direct data acquisition, thereby facilitating more efficient model training \citep{lakshminarayanan2017simple, guo2021calibration}. The uncertainty associated with a prediction is composed of two distinct elements: model uncertainty and data uncertainty. The differences between data and model uncertainties are illustrated in Figure \ref{fig: uncertainty}.

\subsubsection*{Data Uncertainty:}


Data uncertainty, also referred to as aleatoric uncertainty, characterizes the inherent variability within the data itself. In regression problems, this uncertainty, denoted as $\sigma_{d,reg}$, can be inferred directly from the data. This is achieved by adapting the model and loss function to predict a distribution over possible outcomes, rather than a single point estimate. It is managed by adjusting the loss function to capture both the mean and standard deviation for each data point, assuming that the noise follows a normal distribution. Specifically, the model, parameterized by $\theta$,  simultaneously estimates both the mean $\mu_\theta(x)$ and the standard deviation $\sigma_\theta(x)$ of the output distribution $y$. This process involves minimizing the following loss function:
\begin{equation}
\theta^* = \argmin_{\theta} \frac{1}{N} \sum_{i=1}^N \frac{1}{2 \sigma_\theta^2\left(x_i\right)}\left\|y_i-\mu_\theta\left(x_i\right)\right\|_2^2+\frac{1}{2} \log \sigma_\theta^2\left(x_i\right).
\end{equation}
Once trained, the model provides the data uncertainty for a specific input $x$ through the standard deviation of the output estimate, as determined by the learned parameters $\theta^*$. This relationship is mathematically expressed as:
\begin{equation}
    \sigma_{d,reg}(x) = \sigma_{\theta^*}(x). 
\end{equation}

In classification tasks, the predicted output is the probability of a positive or negative classification.
Let $p(y|x)$ represent the probability of a positive outcome and $1-p(y|x)$ the probability of a negative outcome at input $x$. For simplicity, these probabilities are denoted as $p$ and $1-p$, respectively. 
The data uncertainty at input $x$ can be quantified using the concept of entropy from information theory. This is expressed as:
$    \sigma_{d, cls}(x) = \mathrm{H}(p(y|x)), 
$
where the entropy $\mathrm{H}(p(y|x))$ is calculated as:
    \begin{equation}
        \mathrm{H}(p(y|x))=-p \log \left(p\right) - (1-p) \log \left(1-p\right)\label{eq_entropy}.
    \end{equation}
Entropy measures the average amount of information contained in a binary random variable. It reaches its maximum when the probability of each outcome is equal (i.e., $p=0.5$), indicating the highest level of data uncertainty. This reflects a state of maximum uncertainty or indecisiveness about the classification outcome, making it a crucial measure in assessing the confidence of predictions in classification tasks. 

\subsubsection*{Model Uncertainty:}
Model uncertainty, also known as epistemic uncertainty, arises from limitations in the model itself. This can be due to an incomplete sample, errors in the training procedure, or an inadequate model structure. In Bayesian neural networks, model parameters, denoted by $\theta$ are treated as distributions rather than fixed point estimates. This allows using the predictive distribution \eqref{eq_approx_posterior} to estimate model uncertainty.
    

In regression tasks, consider $y^{(m)}(x) = \mu_{\theta^{(m)}}(x)$ as the predictive mean of the outcome for input $x$, given a sampled set of parameters $\theta^{(m)}$. The empirical mean of these predictions over $M$ samples at input $x$ is denoted by  $\mu(y|x)$ , calculated as:
    $\mu(y|x)=\frac{1}{M} \sum_{m=1}^N y^{(m)}(x).$
The model uncertainty at input $x$ is then estimated by the standard deviation across the mean values from different sampled sets of model parameters. This is expressed as:
    \begin{equation}
        \sigma_{m, reg}(x) = \left(\frac{1}{M} \sum_{m=1}^M\left(y^{(m)}(x)-\mu(y|x)\right)^2\right)^{1/2}. \label{eq:m_uncer_reg}
    \end{equation}
    

In classification tasks, model uncertainty is assessed using the posterior distribution  $p(\theta \mid \mathcal{D}^{(t)}_\text{labeled})$, which is approximated by the $q_{\omega}(\theta)$. Define $\bar{p}(x)$ as the expected predicted probability from the Bayesian neural network, calculated as $\bar{p}(x)=\mathbb{E}_{\theta \sim p(\theta \mid \mathcal{D}^{(t)}_\text{labeled})}[p(y \mid x, \theta)]$. The model uncertainty in classification tasks is then determined using the concept of mutual information,
    \begin{equation}
        \sigma_{m,cls}(x) = \mathrm{H}(\bar{p}(x))-\mathbb{E}_{\theta \sim p(\theta \mid \mathcal{D}^{(t)}_\text{labeled})} \mathrm{H}\left(p(y \mid x, \theta)\right), \label{eq:m_uncer_cls}
    \end{equation}
    where $\mathrm{H}(\cdot)$ denotes the entropy function as introduced in \eqref{eq_entropy}. Mutual information, a concept from probability and information theory, measures the amount of information gained about one random variable by observing another. In the context of model uncertainty in classification tasks, mutual information quantifies the difference between the entropy of the average predicted probability and the expected entropy of the individual predicted probabilities for each sample from the model. This difference highlights the degree of uncertainty in the model's predictions.

Using the quantified data and model uncertainties, we construct an uncertainty score as detailed in Section \ref{sec:active details} and a search score in Section \ref{sec:UCB}. These scores act as essential meta-knowledge that guides the algorithm's recommendations. In Section \ref{sec: human}, we explore the role of human experts in leveraging these scores to make final decisions.

As discussed in \S \ref{sec: meta-knowledge}, this uncertainty represents the meta-knowledge of the algorithm. This meta-knowledge serves two primary purposes. Firstly, it enables the identification of such data and its subsequent integration into the model's training data, potentially enhancing performance in line with active learning concepts \citep{reker2015active}. Secondly, it provides a confidence level for each recommendation to human experts, who can then make more informed decisions in this delegated responsibility \citep{fugener2022cognitive}.

\subsection{Algorithm Improvement and Uncertainty Score} \label{sec:active details}


In this section, we delve into the details of the uncertainty score and its application in enhancing the algorithm with each round. Within our framework, multiple properties are considered simultaneously. The algorithm includes $K$ Bayesian neural networks, represented as $\{\mathcal{C}_{\theta_j}\}_{j=1}^K$. In each round $t$, the algorithm predicts outcomes for unlabeled molecules and computes a model uncertainty vector $\boldsymbol{\sigma}_m(\boldsymbol{x}^u)$ for each molecule  $\boldsymbol{x}^u\in \mathcal{D}^{(t)}_{\text{unlabeled}}$. This vector $\boldsymbol{\sigma}_m(\boldsymbol{x}^u)$ comprises $K$ dimensions, corresponding to each property. The vector includes the model uncertainties discussed previously, with continuous values as specified in \eqref{eq:m_uncer_reg} and binary values as detailed in \eqref{eq:m_uncer_cls}.
We define the uncertainty score $r_{un}(\boldsymbol{x}^u)$ for each molecule $\boldsymbol{x}^u$ as the sum of the $K$ model uncertainties, which is
\begin{equation}
    r_{un}(\boldsymbol{x}^u) = \boldsymbol{1}^\top \boldsymbol{\sigma}_m(\boldsymbol{x}^u) =\sum_{k=1}^K (\boldsymbol{\sigma}_m)_k(\boldsymbol{x}^u),
\end{equation}
where $(\boldsymbol{\sigma}_m)_k(\boldsymbol{x}^u)$ is the model uncertainty for the $k$-th property for molecule $\boldsymbol{x}^u$. 

Then, the algorithm identifies $q$ molecules with the highest uncertainty scores for its initial batch of recommendations. Following the principles of standard active learning, these molecules will enhance the model's performance more rapidly than others.

Our approach aligns with the widely adopted model uncertainty strategy \citep{lewis1995sequential}, which involves querying those unlabeled molecules where the current algorithm's predictions are most uncertain. Data characterized by high model uncertainty are not well understood by the existing model. Therefore, labeling such data and incorporating it into the model's training can significantly improve its performance.
\subsection{Target Molecule Searching and Search Score} \label{sec:UCB}
In this section, we delve into the search score and its significance in targeting molecule discovery. Our framework’s search score draws inspiration from the upper confidence bound (UCB) concept, a prevalent strategy in online learning task optimization \citep{powell2012optimal,bayati2022speed}. An UCB-type algorithm sequentially selects promising targets based on the estimated mean $\boldsymbol{\mu}$ supplemented by a term proportional to the standard deviation, expressed as $\beta_t\boldsymbol{\sigma}$. This method of searching, prioritizing the highest mean, maximizes the chances of finding target molecules, while the inclusion of the uncertainty term ensures a balance between exploration and exploitation. 

In our framework, for each round and each unlabeled molecule $\boldsymbol{x}^u\in \mathcal{D}_{\text{unlabeled}}$, we construct an upper confidence bound vector $\text{UCB}(\boldsymbol{x}^u)\in \mathcal{R}^K$, i.e.,
\begin{equation}
    \bold{\text{UCB}}(\boldsymbol{x}^u) = \boldsymbol{\mu}(\boldsymbol{x}^u)+\beta_{t}^{1 / 2} \boldsymbol{\sigma}_d(\boldsymbol{x}^u).
\end{equation}
Unlike conventional UCB, where the standard deviation $\boldsymbol{\sigma}(\boldsymbol{x}^u)$ is used, our approach employs data uncertainty $\boldsymbol{\sigma}_d(\boldsymbol{x}^u)$. The relationship between uncertainty and standard deviation in the model and data is given by:
\begin{equation}
    \boldsymbol{\sigma}^2(\boldsymbol{x}^u) = \boldsymbol{\sigma}_d^2(\boldsymbol{x}^u) + \boldsymbol{\sigma}_m^2(\boldsymbol{x}^u).
\end{equation}
We opt for data uncertainty over standard deviation because model uncertainty arises from inaccuracies in model predictions, rather than reflecting the molecule's potential. Our focus on model uncertainties is specifically on enhancing the model. Thus, our UCB version more accurately captures the potential of the molecule, which is a fundamental difference from traditional approaches. Additionally, $\beta_t$ is progressively reduced to enhance algorithm stability in the later stages of the discovery process.

For multi-objective tasks, we use the aggregate of each UCB score as the search score $r_{se} (\boldsymbol{x}^u)$,  
\begin{equation}
    r_{se} (\boldsymbol{x}^u)=\boldsymbol{1}^\top \bold{\text{UCB}}(\boldsymbol{x}^u)=\sum_{k=1}^K \text{UCB}_k(\boldsymbol{x}^u),
\end{equation}
where $\text{UCB}_k(\boldsymbol{x}^u)$ is the UCB score for the $k$-th property. The algorithm then recommends $h$ molecules with the highest search scores as its second set of recommendations.

\subsection{The Role of Human}
\label{sec: human}


In each round $t$, the algorithm recommends two batches of molecules for human expert review. The first batch contains $q$ molecules $\{\boldsymbol{x}_{un,i}^{(t)}\}_{i=1}^q$ with the highest uncertainty scores $r_{un}$. 
These molecules have the potential to enhance model performance, a topic elaborated on in \S \ref{sec:active details}. 
The second batch consists of  $h$ molecules $\{\boldsymbol{x}_{se, i}^{(t)}\}_{i=1}^h$ with the highest search scores $r_{se}$, likely to be prospective target molecules according to the current algorithm's predictions, as discussed in \S \ref{sec:UCB}. 
There might be overlaps between these batches, resulting in the human experts reviewing a total number of molecules ranging from $\max\{q, h\}$ to $q+h$. Detailed scores accompany these molecules, providing meta-knowledge that guides the experts in deciding whether to follow the algorithm's recommendations or to deviate. Experts are also briefed on the rationale behind these recommendations.

Based on the recommendations and scores, we enable human experts to assess and refine algorithmic suggestions, making final decisions on the $B$ molecules to synthesize and test in this round. Their expertise is crucial for identifying viable structures and discarding impractical suggestions. Particularly in the early stages, where algorithms may lack accuracy, the role of human expertise is crucial. This interaction not only steers the algorithm's search process but also enriches it with positive labels to accelerate learning. Additionally, human experts consider both the remaining budget and the algorithm’s performance. Early rounds might prioritize enhancing the algorithm’s accuracy, while later rounds could allocate more resources to the discovery of promising molecules.

To summarize, humans play three key roles:
    i) They conduct laboratory experiments to synthesize and test molecular properties.
    ii) They utilize domain knowledge, integrating it with algorithm recommendations and meta-knowledge to identify and correct algorithmic errors, guiding the search for potential molecules.
    iii) They manage resources to balance model enhancement and drug discovery efforts, especially as the algorithm evolves.


\section{Experiments} \label{sec: numerical experiements}


To evaluate the performance of our method, we use real therapeutic peptide data from the SATPdb database\footnote{SATPdb is a database of structurally annotated therapeutic peptides: \href{https://webs.iiitd.edu.in/raghava/satpdb/}{https://webs.iiitd.edu.in/raghava/satpdb/}.} for the discovery of therapeutic peptides satisfying multiple properties. The dataset comprises 17,392 unique therapeutic peptides, each experimentally validated for 10 major properties such as anti-hypertensive, anti-cancer, anti-bacterial, and others. 

\subsection{Dataset and Experimental Procedures} \label{sec: exp}
For the experiment, we randomly selected 14,000 peptides from the dataset as our search pool, while the remaining 3,392 peptides formed the test set to evaluate the  models' performance in each round and monitor their improvements. Initially, we assumed no training data, i.e., $|\mathcal{D}^{(1)}_\text{labeled}| = 0$ and $|\mathcal{D}^{(1)}_\text{unlabeled}| = 14,000$. The experimental budget for each round is $B= 50$, and there are a total of $R=50$ rounds for the entire process. This means we need to identify target peptides within $2,500$ experimental tests.  We engaged a domain expert in therapeutic peptides to provide insights and informed her about our algorithm's workings.


At the start of the $t^\text{th}$ round, the models $\{\mathcal{C}^{(t)}_{\theta_i}\}_{i=1}^K$ predict outcomes for each molecule $\boldsymbol{x}^u$ in $\mathcal{D}^{(t)}_\text{unlabeled}$, and compute both uncertainty scores and search scores. The top $q=50$ molecules with the highest uncertainty scores and $h=50$ molecules with the highest search scores, along with their property-specific scores, are presented to the expert for review. The total number of recommendations received by the expert each round ranges from 50 to 100, considering potential overlaps between the two batches. The expert then thoroughly evaluates these molecules and their scores. 
Next, the expert applies her domain knowledge to select 50 peptides $\{\boldsymbol{x}_{i}^{(t)}\}_{i=1}^{50}$ for laboratory experiments and obtain their labels for all $K$ properties $\{\boldsymbol{y}_{i}^{(t)}\}_{i=1}^{50}$.

Subsequently, we update the labeled dataset  $\mathcal{D}_{\text{labeled}}^{(t+1)}=\mathcal{D}_{\text{labeled}}^{(t)}\bigcup\{(\boldsymbol{x}_{i}^{(t)},\boldsymbol{y}_{i}^{(t)})\}_{i=1}^{50}$. This newly labeled dataset $\mathcal{D}_{\text{labeled}}^{(t+1)}$ is then used to refine the algorithm $\{\mathcal{C}^{(t+1)}_{\theta_i}\}_{i=1}^K$ for the next round. This cycle continues for $R=50$ rounds. We undertook three tasks of varying difficulty:
\begin{enumerate}
    \item[] \textbf{Task 1}. Discovery of molecules with anti-cancer, anti-bacterial, and anti-fungal properties, including 148 target molecules in the 14,000-molecule search pool. 
    \item[]\textbf{Task 2}. Discovery of molecules with anti-microbial, anti-viral, and anti-bacterial properties, including 109 target molecules in the 14,000-molecule search pool. 
    \item[] \textbf{Task 3}. Discovery of molecules with anti-bacterial, anti-viral, anti-microbial, anti-cancer, anti-parasitic, and anti-fungal properties, with 33 target molecules in the 14,000-molecule search pool.  
\end{enumerate}

Tasks 1 and 2 involve finding molecules that simultaneously satisfy three properties, while Task 3 requires identifying molecules meeting all six properties. In Task 1, approximately 1\% of the molecules are target molecules, while only $7.8 \text{\textperthousand}$ and $2.4\text{\textperthousand}$ of molecules in Tasks 2 and 3, respectively, meet all required properties. Each task poses a significant challenge in practice.


\subsubsection*{Algorithm Details.}
In our experiments, we set the embedding space dimension at $|\mathcal{Z}|=1,280$ for the ESM2 model discussed in \S \ref{sec:ESM2} for embedding. Each Bayesian neural network model, as outlined in \S  \ref{sec:bayesian_NN}, was implemented with the following specifications: 
\begin{itemize}
    \item Input layer size of 1,280, matching the ESM2 encoding dimension of the peptide molecules;
    \item Six hidden layers with node sizes of 128, 64, 32, 16, 8, and 2, using Relu activation functions ($Relu(\cdot) = \max(0,\cdot)$);
    \item Output layer comprising two nodes and a Softmax layer\footnote{$\text{Softmax}(h_1,h_2)= \left(\frac{e^{\beta h_1}}{e^{\beta h_1}+e^{\beta h_2}},\frac{e^{\beta h_2}}{e^{\beta h_1}+e^{\beta h_2}}\right)$, where $h_1,h_2$ are two inputs from the hidden layer. These inputs are normalized to yield the predicted probabilities for the two classes.}, indicating the probabilities of the input samples being classified as positive or negative.
\end{itemize}

Furthermore, we assumed a normal distribution with mean 0 and standard deviation 0.01, i.e., $\mathcal{N}(0,0.01)$, as the prior for each neural network parameter, updating these parameters using the variational inference method detailed in \S \ref{sec:bayesian_NN}. We employed batch gradient descent for training, with a learning rate of $5\times 10^{-4}$ using the Adam optimizer, a training batch size of 50, and 200 training epochs. All experiments were conducted using three NVIDIA RTX A6000 GPUs and PyTorch.

\subsubsection*{Benchmarks.} \label{sec: benchmark}

The primary performance metric for sequential experiments in drug discovery is the final hit rate,  $\frac{S_\pi}{B\times R}$, reflecting the proportion of target molecules identified among all tested molecules. Another key metric is the recall rate, $\frac{S_\pi^t}{S_\star}$. The recall rate represents the ratio of the total number of target molecules discovered in the first $t$ rounds ($S_\pi^t$) to the overall number of target molecules in the search pool ($S_\star$). It should be noted that in real-life drug discovery, the recall rate is unknown since the actual number of target molecules, $S_\star$, is not disclosed.  However, for this experiment, it serves as a useful metric for retrospective algorithm comparison. 

In our main evaluation, we compared  our approach with five benchmark policies: 
\begin{enumerate}
    \item \textit{Random}: Selects molecules randomly for laboratory tests.
    
    \item \textit{Search-based (Pure search)}: Selects molecules based on predicted mean values.
    
    \item \textit{Uncertainty-based (Active learning)}: Selects molecules using model uncertainty.
    
    \item  \textit{Upper Confidence Bound (UCB)}: Selects molecules based on UCB scores.

    \item  \textit{Train-then-search}: Initially trains a machine learning model with collected samples and then selects molecules using UCB search scores.
    \begin{enumerate}
        \item  \textit{Random-then-UCB}: Initially selects molecules randomly for training, then uses UCB search scores for selection.
        \item \textit{Active learning-then-UCB}: Begins with active learning for model training, then uses UCB search scores for selection.
    \end{enumerate}
\end{enumerate}

The two train-then-search strategies demonstrate how combining uncertainty-based and search-based approaches performs without human involvement. As presented in \S \ref{sec: results}, our method outperforms all benchmark policies by at least 100\%.




\subsection{Results Compared to Benchmark Strategies}
\label{sec: results}

In this section, we describe the evaluation results of the three tasks. 
We recorded the cumulative recall rate ($\frac{S_\pi^t}{S_\star}$) for each round $t$ and the final hit rates for our and the benchmark policies across the three tasks. For each policy, we repeated the experiment 20 times with random initial data and calculated the mean and the confidence interval, using two standard deviations for each round. 

Table \ref{tab:result} displays the average and standard deviation of the final hit rates $\frac{S_\pi}{B\times R}$ for our policy and the benchmark policies over 20 repetitions in three drug discovery tasks. In each task, our method significantly outperformed all benchmarks. According to the final hit rates shown in the table, our approach achieved at least three times the performance of the random exploration baseline in every task. This baseline refers to mass synthesizing and testing possible molecules through high-throughput laboratory techniques in practice without human advisory or algorithmic assistance. Moreover, our results nearly doubled those of the second-best method, the AL-UCB policy, which involves no human involvement. This aligns with existing literature that emphasizes the crucial role of human expertise in drug discovery using ML \citep{sturm2021coordinating} and highlights the value of human-specific information in tackling challenging tasks \citep{balakrishnan2022improving}.


\begin{table}[htbp]
\TABLE
{Final Hit Rates of the Three Tasks\label{tab:result}}
{\setlength{\tabcolsep}{10pt} 
\renewcommand{\arraystretch}{1.2} 
\begin{tabular}{c | c c c c c c c}
    \hline
        &Ours & RD-UCB & RD & Pure Search & AL-UCB &  AL & UCB\\
    \hline
        \multirow{2}{*}{Task 1} &   \textbf{2.97\%} & 1.61\% & 1.02\% &  1.29\% & 1.65\% &  1.16\% &  1.69\%  \\
              & (0.25\%) & (0.13\%) & (0.23\%) & (0.17\%) & (0.31\%) & (0.14)\% & (0.28\%) \\
        \multirow{2}{*}{Task 2} & \textbf{2.57\%} & 1.30\% & 0.68\% & 0.95\% & 1.59\% & 0.97\% & 1.27\%\\
              & (0.26\%) &(0.22\%) & (0.18\%) & (0.20\%) & (0.21\%) & (0.06\%) & (0.16\%) \\
        \multirow{2}{*}{Task 3} & \textbf{0.76\%} & 0.34\% & 0.14\% & 0.31\% & 0.46\% & 0.30\% & 0.44\%\\
              & (0.07\%) &(0.13\%) & (0.05\%) & (0.14\%) & (0.03\%) & (0.16\%) & (0.03\%) \\
    \hline
\end{tabular}}
{\textit{Note}. This table displays the final hit rates of our approach compared to benchmarks for the three tasks. Our approach nearly doubled the results of the second-best method. This advantage was even more pronounced in the more challenging tasks, specifically Task 3 in our experiments. RD: random, AL: active learning, UCB: upper confidence bound.}
\end{table}

\begin{figure}[htp!]
\FIGURE
{\includegraphics[scale=0.3, trim={0 2cm 0 0},clip]{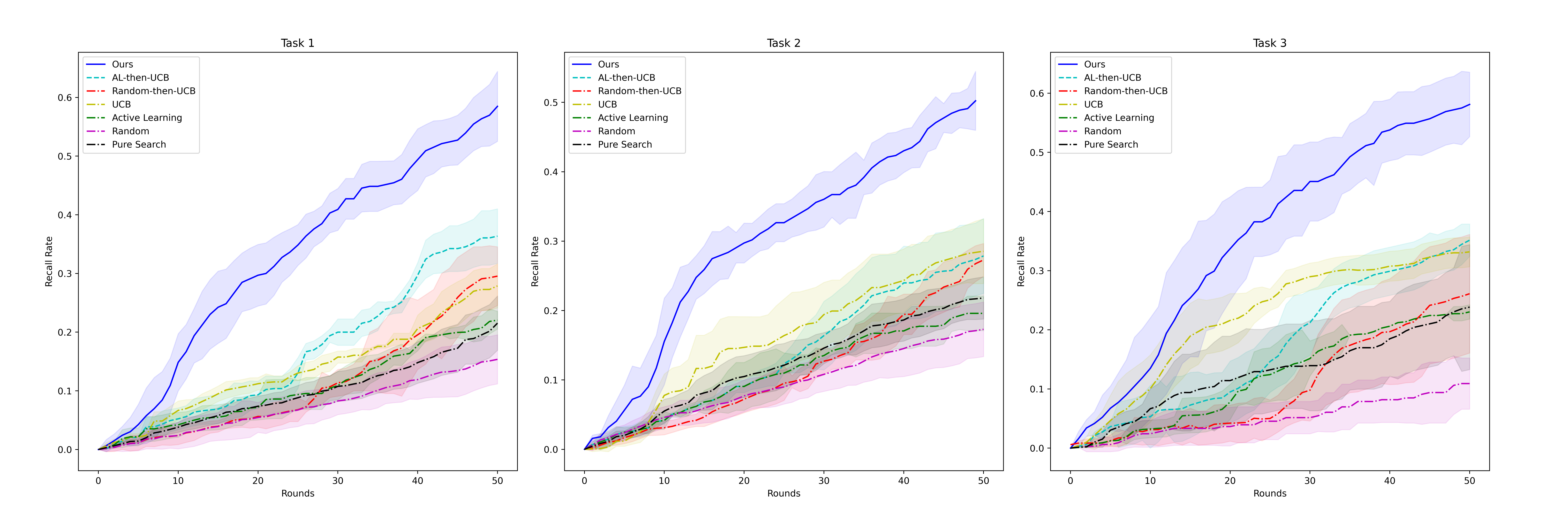}}
{Results of Different Policies Searching for Target Drug Molecules.\label{fig: exp result}}
{This figure shows the cumulative recall with an error bar indicating the largest deviation for each round in the three experimental results in discovering therapeutic peptides. The dark blue lines represent the performance of our human-in-the-loop approach.
}
\end{figure}


To further examine the performance of all methods at each round, we present in Figure \ref{fig: exp result} the accumulative recall rate for each round across all three tasks. We found that our method, consistent with the results shown in Table \ref{tab:result}, consistently achieved the highest final recall rate across all tasks. This advantage becomes even more pronounced in the more challenging tasks. Additionally, we observed that our algorithm is quicker at identifying target molecules initially, becoming most evident between 20-30 rounds. The effectiveness of our approach is derived from the improved accuracy of models through collaborative decision-making between the algorithm and human experts initially, when algorithms possess limited knowledge. This further leads to the recommendation of higher-quality molecules for screening, creating a beneficial cycle of continuous improvement.

\subsection{Ablation Analysis of the Algorithm} 

We analyzed the impact of each component of the algorithm independently to discern the contributions and effects of individual elements. Our findings reveal that exclusively using active learning, molecule search, and a combination of active learning followed by UCB, resulted in notably inferior outcomes, as detailed in the last three columns of Table \ref{tab:result}. It became evident that neither of the components (AL and UCB) nor their straightforward combination could match the performance of our comprehensive framework.

\begin{figure}[htbp]
\FIGURE
{\includegraphics[scale=0.45,trim={0 2cm 0 0.5cm},clip]{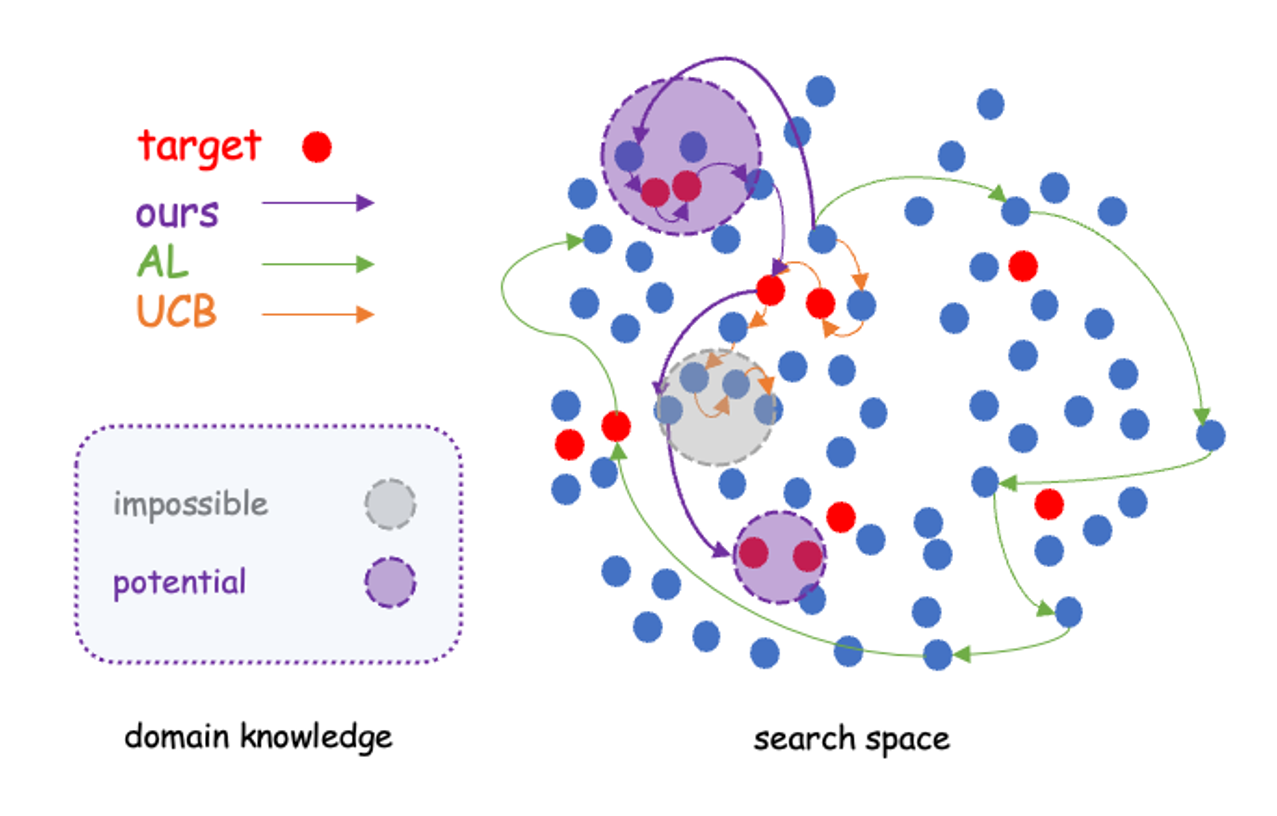}}
{Search Patterns of Different Policies.\label{fig: result analysis searching}}
{This figure shows the search patterns of the three policies in Task 2. Active learning (AL) predominantly focuses on molecules with unexplored structures, whereas UCB tends to search within a local region. In contrast, our human-in-the-loop approach effectively balances exploration and exploitation. }
\end{figure}

To illustrate the impact of human involvement on model performance, we first analyzed the search patterns during the training phase, as depicted in Figure \ref{fig: result analysis searching}. We observed that UCB generally focuses on areas with some level of exploration, guided by uncertainty. However, given the extensive search space and the rarity of target molecules, UCB often becomes restricted to regions where target molecules were initially discovered. This tendency likely explains why UCB is effective in early rounds but less so in later ones. In contrast, active learning (AL) consistently targets unexplored regions, choosing points distant from previously identified molecules. This strategy enables quicker development of an effective model but is less effective in searching for target molecules.
\begin{figure}[htbp]
\FIGURE
{\includegraphics[scale=0.45,trim={0 2cm 0 3cm},clip]{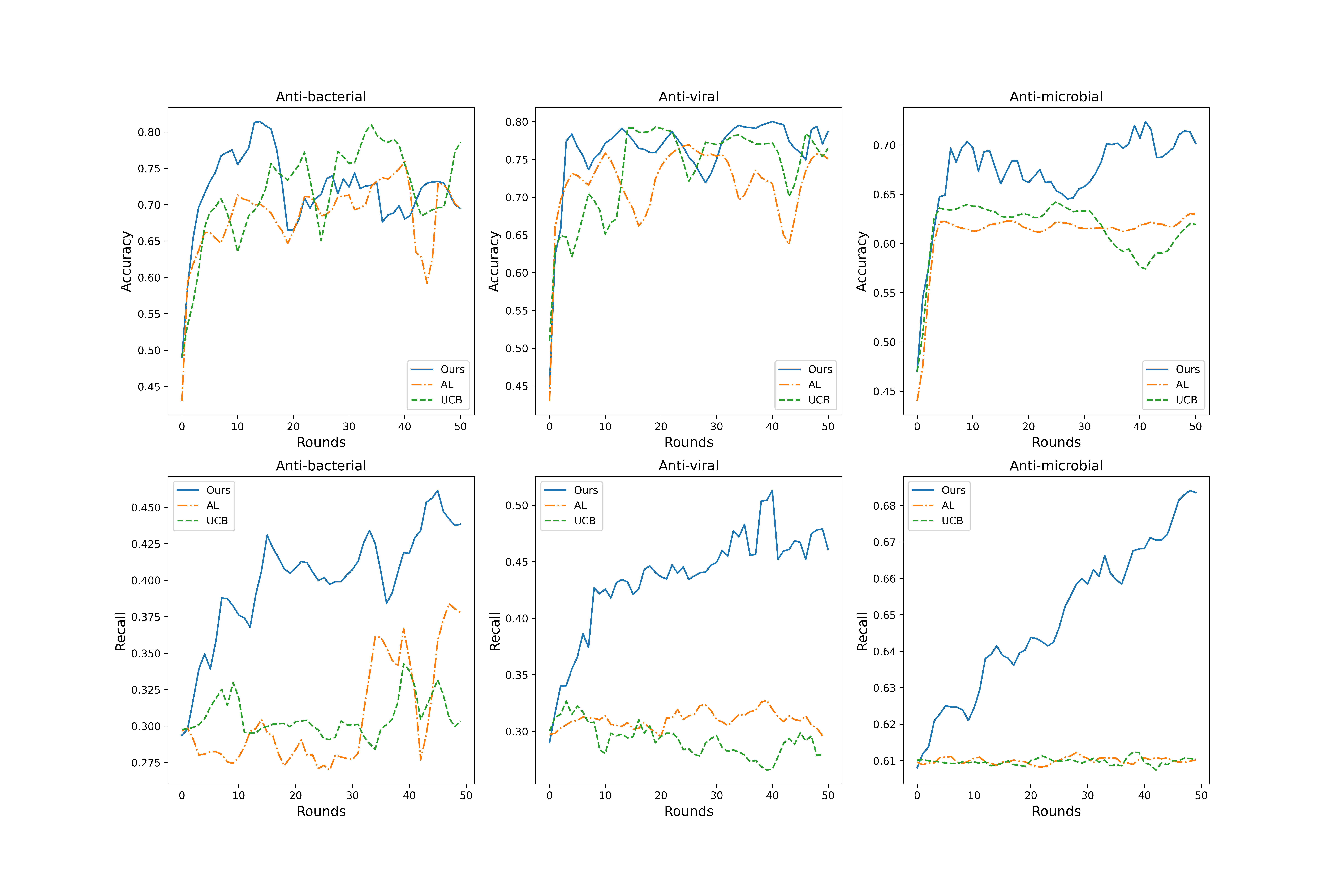}}
{Tracking Algorithm Performance for Task 2.\label{fig: result analysis recall and accuracy}}
{This figure traces the evolution of prediction accuracy and recall on the validation set for the three classifiers in Task 2, each trained using our approach, UCB, and active learning, respectively. Our method markedly enhances the algorithm's recall compared to the exclusive use of UCB or AL, particularly in the initial phases of the sequence experiments.}
\end{figure}

We also monitored the prediction accuracy and recall of the algorithm for the three properties in Task 2, using a test set comprising 3,392 data points that were initially left out. The results are presented in Figure \ref{fig: result analysis recall and accuracy}, with accuracies for the three properties displayed in the first row and recalls in the second row. While the accuracy of our algorithm does not show significant improvement, the recall shows a significant increase.\footnote{We note that, with a slight misuse of terms, this recall refers to the algorithm's recall for a classification task, which is different from the recall rate we defined earlier as the ratio of the total number of target molecules discovered to the total number of target molecules in the search pool, $\frac{S_\pi^t}{S_\star}$.} 

\subsubsection*{Discussion.}
We identify two factors contributing to the performance of our framework.

The first factor is the \textit{warm start-up} of the algorithm when dealing with an imbalanced dataset. From the beginning, our method consistently outperforms all benchmark methods, including those focused solely on identifying good molecules, as depicted in Figure \ref{fig: exp result}. This improvement is attributed to the accelerated learning of the algorithm when human experts assist in the search for target molecules, thereby enhancing its ability to accurately predict positive labels. We note that this is crucial for classification algorithms dealing with imbalanced data, which is often the case in drug discovery, where positive labels are very rare. Typically, algorithms struggle to train effectively without a sufficient number of positive samples. In contrast, human experts can provide valuable guidance for a warm start-up. As the rounds progress, the algorithm, benefiting from improved training, suggests more promising candidates. Thus, the synergistic collaboration between the algorithm and human experts enables our method to rapidly identify target molecules.

The second factor is the effective \textit{balance between model improvement and target search}, guided by human expert knowledge. In our approach, we integrate model optimization and molecule discovery, leveraging the strengths of search algorithms under uncertainty, active learning, and human private information. Human experts possess knowledge about specific structures that may indicate desired properties, enabling them to guide the algorithm toward potential target molecules. Additionally, they can identify and avoid molecules that are likely unsuitable (illustrated in grey in Figure \ref{fig: result analysis searching}), thereby conserving experimental resources. 


\subsection{Achieving Complementarity Between Human and Algorithm}
\begin{figure}[htbp]
\FIGURE
{\includegraphics[scale=0.4, trim={0 0 0 1cm},clip]{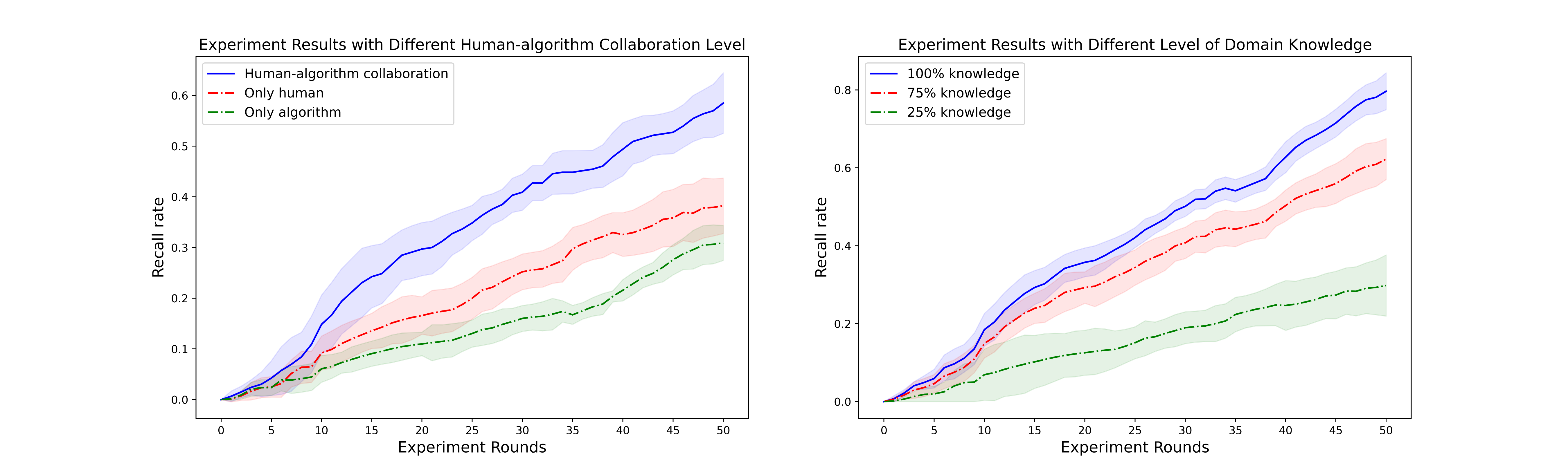}}
{Complementarity of Human-Algorithm Collaboration and the Impact of Human Knowledge Level. \label{fig: add exp result}}
{The left figure illustrates the cumulative recall rate for each round across three experimental settings, each with varying degrees of human-algorithm collaboration. The right figure shows the cumulative recall rate for each round, focusing on the algorithm's interaction with humans possessing different levels of domain knowledge. We observe that collaboration leads to a marked improvement. Furthermore, greater human expertise results in a higher recall rate and a reduced standard deviation. }
\end{figure}

Starting from this section, we focus more on the synergistic effect of human and algorithm collaboration. Specifically, in this section, we are interested in understanding the complementarity between human and algorithm. We aim to determine the extent to which collaboration between human and algorithm can outperform the efforts of only humans or algorithms in the drug discovery process. We compare the results under the following three settings:
\begin{itemize}
    \item \textit{Human-algorithm collaboration}: The human expert considers the algorithm's suggestions and makes final decisions based on her domain knowledge.
    \item \textit{Human only}: The human expert relies exclusively on her own knowledge, disregarding the algorithm's outputs and suggestions.
    \item \textit{Algorithm only}: The human expert selects the molecules the algorithm suggests.
\end{itemize}
We conduct identical experiments for Task 1 as outlined in Section \ref{sec: exp}. For each scenario, we perform the target peptide searching experiments 20 times, using identical parameters. We compare the performance of these three settings and present the experimental results in the left figure of Figure \ref{fig: add exp result}. The results indicate that the collaboration between human and algorithmic input achieves the best results in terms of recall rate, compared to decisions made solely by humans or algorithms. Specifically, performance increased by about 50\%. This demonstrates significant complementarity between human and algorithm, where algorithms offer a data-driven foundation, and humans apply a contextual overlay to refine and implement these insights effectively \citep{fugener2021will, fugener2022cognitive, donahue2022human}. 

Especially in sequential experiments, human-algorithm collaboration creates iterative feedback loops, in which the collaboration facilitates a continuous exchange of information where human inputs help to refine and adjust algorithmic outputs. As in our context,  scientists can provide feedback on the algorithm's molecule selection, which can be used to train the algorithm further, enhancing its accuracy. This iterative loop not only improves the algorithm over time but ensures that the algorithm remains aligned with changing human priorities and external conditions.

On one hand, algorithms can help mitigate some of the cognitive biases that humans are prone to, such as confirmation bias or overconfidence. By presenting data-driven insights, algorithms encourage humans to reconsider assumptions and explore new lines of inquiry that might otherwise be overlooked. On the other hand, exposure to algorithmic decision-making can also educate and inform human decision-makers, leading to improved judgment and expertise over time. This adaptive learning process is crucial for maintaining the relevance and effectiveness of decision-making systems under uncertainty in which human decision-makers do not have full knowledge of the environment.

Interestingly, reliance on human experts alone outperformed sole dependence on the algorithm, highlighting the value of human experts' private knowledge \citep{balakrishnan2022improving}. This outcome underscores the critical role of private domain knowledge in enhancing the efficiency and effectiveness of molecule screening in drug discovery. Furthermore, in more challenging tasks where human expertise is crucial, our human-in-the-loop framework is expected to provide increased use in assisting experts in identifying target molecules.

\subsection{Impact of Human Knowledge Levels}
We hypothesize that the level of human knowledge may also have an impact on the final performance. In this section, we analyze the performance of our framework as the algorithm interacts with human experts who possess varying levels of domain knowledge. We evaluate the performance in scenarios where the expert has high-level, medium-level, and low-level domain knowledge about therapeutic peptides. 

Since evaluating the knowledge level in practice is challenging, we simulate the results. Specifically, we enlist a volunteer with minimal domain knowledge to act as a human expert, thereby minimizing the impact of prior knowledge on the experimental results. During the experiments, for each peptide molecule, we disclosed the true labels to the participant with probabilities of $p=25\%, 75\%, 100\%$, without informing them of the actual percentage. This approach follows the work of \cite{fugener2022cognitive}, which suggests that humans might not be able to accurately judge their abilities for difficult tasks. We repeated Task 1 from \S \ref{sec: exp} 20 times for each level of knowledge, maintaining all other parameters identical to those in \S \ref{sec: results}. We then compared the recall rate in experiments with varying levels of expert domain knowledge and displayed the results in the second figure of Figure \ref{fig: add exp result}.

The findings from the figure reveal that enhanced knowledge contributes to superior performance, characterized by a higher recall rate and a reduced standard deviation. Specifically, when 100\% knowledge is disclosed, we achieve an 0.8 recall rate by the end of 50 rounds. We note that this recall rate is not 100\% for two reasons. First, the volunteer is unaware of the actual percentage of knowledge they receive. Second, our framework operates under a procedure where algorithms recommend two lists of molecules, and the volunteer makes the final decision from these two lists. Their ability is limited in that they cannot exclusively rely on this given knowledge but must collaborate with the algorithm in adhering to our framework.

When 75\% and 25\% knowledge levels are disclosed, the final recall rates are around 0.6 and 0.3, respectively. It is interesting to note that with 25\% knowledge, the final recall rate of 0.3 is nearly the same as when the algorithm alone makes the decision. The phenomenon of minimal or no improvement in the collaboration between human and algorithm has also been observed \citep{lebovitz2021ai, balakrishnan2022improving}. It suggests that insufficient domain knowledge might render human contributions ineffective or even detrimental. This aligns with prior research indicating that under certain conditions, uninformed human intervention can introduce noise or bias and decrease the reliability of algorithmic processes \citep{shin2024misinformation,fugener2021will,fugener2022cognitive}. This observation emphasizes the need for appropriate training of humans to maximize the potential of human-AI collaboration. It suggests that AI systems should be designed with interfaces and feedback mechanisms that are intelligible to users with varying levels of expertise, facilitating better integration of human judgment.


\subsection{Value of Meta-knowledge}

In addressing complex challenges, human experts contribute private or domain-specific knowledge, whereas algorithms derive knowledge from data. Beyond the foundational knowledge required, meta-knowledge plays a pivotal role in enhancing complementarity during collaboration. Meta-knowledge is formally defined as the ability to evaluate one's own capabilities.
Algorithms possess a high level of meta-knowledge, enabling them to assess their capabilities accurately and share this assessment with human experts. In contrast, humans often find it challenging to articulate the decision-making rules they utilize and may inaccurately assess their own skills \citep{fugener2021will, fugener2022cognitive}. This experiment aims to highlight the significant value of the meta-knowledge that algorithms can offer in drug discovery tasks.

A primary reason for using a Bayesian neural network in our study is its nature as a probabilistic model. Unlike conventional models, it not only generates predictions but also estimates the distribution of these predictions, allowing for the calculation of uncertainties, as detailed in \S \ref{sec:bayesian_NN} and \S \ref{sec:uncertainty_quant}. This ability to quantify uncertainties and search scores represents vital meta-knowledge, enhancing collaboration between humans and the algorithm.

\begin{figure}[htbp]
\FIGURE
{\includegraphics[scale=0.4]{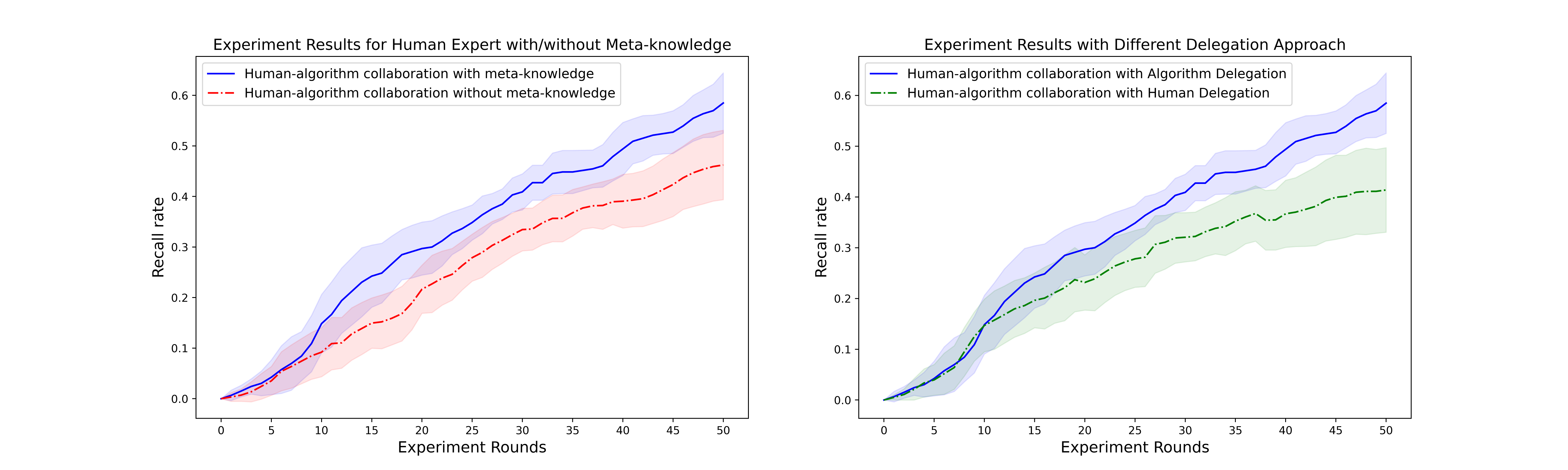}}
{Impact of Meta-knowledge and Delegation of Work \label{fig: add exp result2}}
{The left figure illustrates the cumulative recall rate with and without meta-knowledge presented to humans. Our findings indicate that in the absence of meta-knowledge from the algorithm, learning significantly slows down, with performance deteriorating by more than 40\% compared to scenarios where meta-knowledge is provided. The right figure shows the cumulative recall rate for various work delegation strategies. Our findings suggest that delegation from algorithms to humans improves performance by more than 50\% compared to delegation from humans to algorithms.}
\end{figure}

To assess the impact of this meta-knowledge, we designed an experiment comparing our framework's performance with and without providing this information to the human experts. In the scenario lacking meta-knowledge, we present only the recommended molecules to the human experts, omitting the uncertainty and search scores. Consequently,  human experts must rely solely on their domain-specific knowledge to make selections from these recommendations.

We present the results in the first figure, as shown in Figure \ref{fig: add exp result2}. Our findings indicate that in the absence of meta-knowledge from the algorithm, learning significantly slows down, with performance deteriorating by more than 40\% compared to scenarios where meta-knowledge is provided. Furthermore, the performance only marginally surpasses that of humans working independently, as demonstrated in Figure \ref{fig: add exp result}. This highlights the critical role of the algorithm's meta-knowledge in this task,  illustrating the extent to which human experts rely on AI not just for its recommendation but for an understanding of the reliability and relevance of the recommendation.

\subsection{Delegation of Work}

In our final experiment, we assess the efficacy of various work delegation strategies. The collaboration between humans and algorithms features two principal delegation methods: \textit{from humans to algorithms} and \textit{from algorithms to humans} \citep{fugener2022cognitive}. Our human-in-the-loop framework exemplifies algorithm-to-human delegation, where algorithms perform preliminary screenings and then pass the responsibility for final decision-making to the humans. Conversely, in traditional algorithm-assisted drug discovery approaches, delegation flows from humans to algorithms; humans select molecules for screening and depend on algorithmic predictions to identify the most promising candidates. While this method benefits from initial human insight, it might limit the search space too early in the process, potentially overlooking promising candidates that an algorithm might have identified in a broader scan.

We conducted an additional experiment where humans initially selected 100 molecules, and then the algorithms were tasked with choosing 50 out of these 100 molecules based on high prediction values. The outcomes of this experiment are depicted in the second figure of Figure \ref{fig: add exp result2}. Our findings suggest that, within our context, delegation from algorithms to humans improves performance by more than 50\% compared to delegation from humans to algorithms.  This substantial improvement is largely attributable to the extensive search space inherent in drug discovery tasks. Algorithms excel in initial screenings, while human experts adeptly integrate their private and domain-specific knowledge with algorithmic recommendations, leading to more effective decision-making.

\section{Discussion and Concluding Remarks} \label{sec: conclusion}

We conclude with a few remarks on the rationale and novelty of our framework. Central to our approach is the integration of deep learning algorithms with the domain-specific expertise of human experts. While algorithms are adept at rapid data processing and pattern recognition, they often face challenges in the data-scarce environments typical of drug discovery. Conversely, human experts, with their domain knowledge, effectively guide the initial stages of algorithmic exploration but are limited by the immense volume of molecular data and may miss complex patterns that algorithms can detect. 

This synergy between human intuition and algorithmic precision forms the cornerstone of our method, enhancing the efficiency and efficacy of the drug discovery process. Our approach enables humans to make final decisions and override algorithmic recommendations, thus achieving complementarity, creating the strongest incentive for the adoption of such a combined human-AI system. This is in line with findings from \cite{dietvorst2018overcoming} that people prefer algorithms they can adjust, even minimally. 


Our framework is developed by synergizing Bayesian uncertainty estimations, Bayesian neural networks, and the use of cutting-edge transformer deep learning models. The use of a Bayesian neural network highlights the unique capability of certain AI models to not only provide predictions but also articulate the uncertainty associated with these predictions. Humans often struggle to assess and articulate their own knowledge and decision-making processes, which can lead to overconfidence or undue hesitancy in their judgments. By providing a quantifiable and unbiased assessment of uncertainties (i.e. meta-knowledge), the algorithm's uncertainty quantification compensates for this human limitation, enhancing the overall decision-making accuracy. It not only exceeds the capabilities of traditional methods but also surpasses those that depend solely on human expertise.

The research also provides several managerial implications. First, it shows that collaboration between human experts and algorithms leads to substantial improvements in drug discovery. This underscores an essential insight for management: the optimal use of AI in complex decision-making environments should augment, not replace, human expertise. Managers should consider strategies to foster effective human-algorithm partnerships, which could include training programs to enhance staff AI literacy, developing interfaces that facilitate easier interpretation of AI outputs, and creating feedback mechanisms for continual refinement of both human and algorithmic inputs.

Second, this research highlights the enduring value of human intuition and domain-specific knowledge, particularly in scenarios where algorithms may lack contextual understanding. Moving forward, managers should strive to balance AI integration with the cultivation of deep domain expertise, ensuring that technological advancements enhance decision-making processes without undermining the unique strengths of human judgment.

Third, the significant performance gap between 25\% and 75\% knowledge disclosure points to a threshold effect in the utility of knowledge. Below this threshold, human inputs fail to significantly enhance algorithmic outputs, potentially because the partial knowledge is insufficient for making informed corrections to algorithmic errors or oversights. Hence, organizations should tailor training programs to the varying expertise levels of employees to maximize the benefits of human-AI collaboration. By ensuring that all team members possess at least the threshold level of knowledge about the specific tasks and AI tools in use, companies can enhance the overall effectiveness of these collaborations.

Finally, the dramatic decline without the inclusion of the algorithm's meta-knowledge sheds light on the need for AI systems with greater transparency and explainability.  AI systems should be designed to articulate their level of certainty and the factors influencing their decisions transparently.  This transparency and explainability are critical for building trust and understanding between human users and algorithmic systems. Designing AI systems that clearly articulate how decisions are made, what data is used, and the certainty of each prediction can significantly enhance user confidence and reliance on these technologies. This approach involves not just the technical design of algorithms but also the user interface and experience design aspects, which should facilitate easy access to and understanding of meta-knowledge. 

Our works also have some limitations. The experiments are simulated on a dataset of previously studied peptides and designed in collaboration with domain experts, though carefully structured, there is potential for bias due to expert familiarity with certain peptides. This limitation could be addressed in future studies by incorporating a broader range of peptides and involving experts from varied fields to minimize familiarity biases. Additionally,  developing a more sophisticated scoring mechanism that accurately weights these properties according to their significance is a critical area for future research. Our further exploration also focuses on enhancing the delivery of meta-knowledge to experts and examining the dynamics of collaboration between algorithms and diverse groups of human experts in complex tasks.



\ACKNOWLEDGMENT{%
}

%
%
%


\bibliographystyle{informs2014} 
\bibliography{reference}


\end{document}